\documentclass[]{bytedance_seed}

\usepackage[toc,page,header]{appendix}
\usepackage{algorithm}
\usepackage{listings}
\usepackage{url}
\usepackage{tabularx}
\usepackage{amsfonts}
\usepackage{amsmath}
\usepackage{amssymb}
\usepackage{nicefrac}
\usepackage{wrapfig}
\usepackage{needspace}
\microtypesetup{expansion=false}

\makeatletter
\patchcmd{\@bottomtitlebar}{\vskip 6mm}{\vskip 4mm}{}{}
\makeatother
\patchcmd{\mymaketitle}{\tcbset{top=6mm}}{\tcbset{top=4mm}}{}{}
\patchcmd{\mymaketitle}{\tcbset{bottom=6mm}}{\tcbset{bottom=4mm}}{}{}
\patchcmd{\mymaketitle}{\vskip 6mm}{\vskip 4mm}{}{}
\patchcmd{\maketitle}{\vskip 8mm}{\vskip 4mm}{}{}

\newcommand{\twolinehead}[2]{\begin{tabular}[b]{@{}c@{}}#1\\#2\end{tabular}}

\definecolor{codeblue}{RGB}{36,82,160}
\definecolor{codegreen}{RGB}{45,120,70}
\definecolor{codepurple}{RGB}{125,65,150}
\lstdefinestyle{pytorch}{
  language=Python,
  basicstyle=\fontsize{8}{9.5}\selectfont\ttfamily,
  keywordstyle=\color{codeblue}\bfseries,
  commentstyle=\color{codegreen},
  stringstyle=\color{codepurple},
  showstringspaces=false,
  columns=fullflexible,
  keepspaces=true,
  breaklines=true,
  breakatwhitespace=true,
  xleftmargin=1em
}

\title{Hierarchical Residuals: A Minimalist Decoupling of Residual Width from Compute Width}
\author[1,2,*]{Ziteng Wang}
\author[2]{Jianqiao Lu}
\author[2]{Xinyu Ma}
\author[1]{Jintao Zhang}
\addtolist[1]{Jun Zhu}{\authorlist}{\authorformat}{\protect\\[2pt]}
\author[2,\dagger]{Yiyuan Ma}
\author[1,\dagger]{Jianfei Chen}
\affiliation[1]{Tsinghua University}
\affiliation[2]{ByteDance Seed}
\contribution[\dagger]{Corresponding authors.}

\abstract{
Residual connections underpin Transformers, but they implicitly tie the residual
width carried through depth to the compute width used inside each layer.
We introduce \emph{Hierarchical Residuals} (HiRes), a minimalist architectural change that decouples these dimensions,
widening the residual state while keeping per-layer compute width fixed.
HiRes groups layers into blocks: standard $d$-dimensional PreNorm
layers operate inside each block, while a wider $D$-dimensional PostNorm outer
residual connects blocks. Learned read/write projections bridge the two
spaces at block boundaries, leaving individual Transformer layers
unchanged.
Across different model sizes and depths, HiRes consistently outperforms vanilla residual addition,
with fitted scaling laws indicating a $1.30\times$ improvement in FLOPs efficiency.
On OLMoE-1B-7B pretraining for $500$B tokens, HiRes achieves $0.0208$ lower training loss and $1.14$-point higher downstream accuracy.
HiRes is also competitive with other residual expansion methods such as VWN, mHC and AttnRes with less activation memory and I/O costs.
Built from only dense projections and normalizations, it requires neither specialized kernels nor infrastructure-level optimizations.
HiRes thus exposes residual width as a practical scaling variable using only standard operators.
}

\date{\today}

\begin{document}
\maketitle
\begingroup
\renewcommand{\thefootnote}{\fnsymbol{footnote}}
\footnotetext[1]{Work done during Ziteng Wang's internship at ByteDance Seed.}
\endgroup

\section{Introduction}

Residual connections~\citep{he2016deep} are the backbone of modern Transformers. A standard
PreNorm layer reads from a normalized copy of the current hidden state,
computes an update, and adds that back to the original state
~\citep{vaswani2017attention,xiong2020layer}. Repeating this operation yields
a simple yet effective transport rule: one $d$-dimensional state is carried
through depth along an identity path, while every attention and feed-forward
module contributes an additive update. This design makes deep optimization
practical, but it also ties the width of the state carried through depth to
the width of the working space of every layer.

\begin{figure}[!t]
  \centering
  \includegraphics[width=\linewidth]{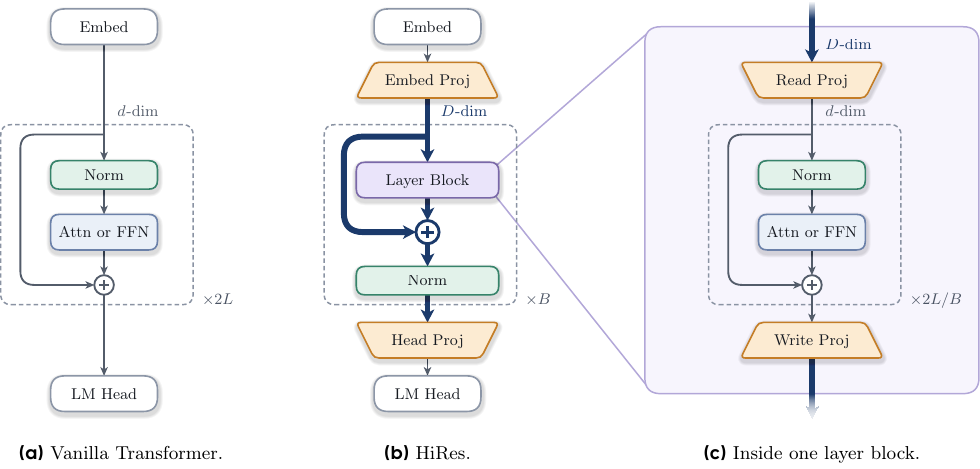}
  \caption{\textbf{HiRes uses a two-level residual hierarchy.}
  (a) Vanilla PreNorm carries one $d$-dimensional residual stream.
  (b) HiRes carries a $D$-dimensional PostNorm outer residual across layer blocks.
  (c) Each block reads $D\!\to\!d$, runs $2L/B$ vanilla sublayers on the
  $d$-dimensional inner residual, and writes back $d\!\to\!D$.
  Gray and navy paths have width $d$ and $D$; orange trapezoids
  are the added dense interfaces, and dashed frames mark repeated residual units.}
  \label{fig:hires-overview}
\end{figure}

Recent residual architectures relax this constraint in different ways.
Alternating Updates (AltUp) maintains multiple residual
streams,
with each layer computing one stream and predicting the remainder~\citep{baykal2023altup}.
Hyper-Connections (HC) likewise uses multiple streams but learns how
layers read from, write to, and combine them with lightweight projectors
~\citep{zhu2025hyperconnections}; Virtual Width Networks (VWN) generalizes
this formulation by splitting each stream into finer granularity~\citep{seed2025vwn};
Manifold-Constrained Hyper-Connections (mHC) stabilizes HC by constraining its
residual mixing operators to a doubly stochastic manifold
~\citep{xie2025mhc}. Attention Residuals instead retrieves
representations from earlier layers using attention over
depth~\citep{kimiteam2026attentionresiduals}. These methods introduce
different residual operators with different update rules. Nevertheless, they share two recurring departures from vanilla
residual addition: (i) the effective residual state is expanded, and (ii) residual accumulation becomes explicitly regulated, keeping the residual scale in check.
This suggests that residual-state capacity is an
architectural variable in its own right, rather than a quantity that must always equal model width.

However, their per-layer expanded states and intricate mixing add activation
traffic and memory-bound residual operators, typically requiring fusion,
recomputation, overlap, or specialized kernels to recover throughput~\citep{xie2025mhc}.
If the benefit stems from the wider state itself rather than particular
operators, a natural question arises: \textbf{can Transformers
benefit from a wider residual state with as little architectural change as possible?}

To this end, we introduce \emph{Hierarchical Residuals} (HiRes). HiRes groups
adjacent Transformer layers into layer blocks and organizes residual
computation at two levels. Inside each block, the model runs ordinary
$d$-dimensional PreNorm Transformer layers. Across blocks, it maintains a
$D$-dimensional outer residual state with the familiar additive update ($D>d$). A
read projection maps the outer state into a block, and after the inner residual accumulation, a
write projection maps the resulting inner state back.
The block write is added to the outer residual and the sum is then
post-normalized, which regulates the residual scale. Thus, the outer recurrence retains the familiar
PostNorm residual, while the inner Transformer layers are unchanged. In this way,
the residual expansion ratio $D/d$ can be selected independently of the
compute width $d$ for each inner layer. Because read/write operations occur only at
block boundaries, the additional parameters and compute costs are amortized over depth.
Figure~\ref{fig:hires-overview} places the two levels side by side: the outer
recurrence is the residual motif of a vanilla PostNorm Transformer, with a whole layer
block in place of a sublayer.

Empirically, HiRes consistently outperforms vanilla residual
addition across model sizes and depths. Under fitted loss-compute scaling
laws, it is $1.30\times$ more FLOP-efficient. As depth scales up at fixed width,
HiRes continues to improve through $128$ layers where the baseline exhibits diminishing returns.
Against Hyper-Connections, mHC, VWN, and Attention Residuals, HiRes remains
comparable in quality at lower activation memory and I/O costs.
On OLMoE-1B-7B~\citep{muennighoff2024olmoe} pretraining for 500B tokens (Section~\ref{sec:long-horizon-moe-pretraining}), HiRes achieves
$0.0208$ lower training loss and $1.14$-point higher downstream accuracy.
Overall, HiRes provides a minimalist way to expand the residual state
without widening per-layer compute.
It shows that decoupling residual width from compute width is beneficial, and that
this decoupling is available with standard operators.

\section{Method}
\label{sec:method}

\subsection{Vanilla Residual Addition}

Consider a decoder-only Transformer with $L$ layers and hidden width $d$.
Each layer contains an attention module and an FFN module, giving $2L$
residual sublayers. Let $x_{i-1}\in\mathbb{R}^d$ denote the residual
state before sublayer $i$. A standard PreNorm Transformer can be written as
\begin{equation}
  x_{i}
  =
  x_{i-1} + F_i\!\left(\operatorname{RMSNorm}_i(x_{i-1})\right),
  \qquad i=1,\ldots,2L,
  \label{eq:vanilla-residual}
\end{equation}
where $F_i:\mathbb{R}^d\rightarrow\mathbb{R}^d$ is an attention or FFN
transformation. Thus the state carried through depth (residual width, dimension of $x_i$) and the state processed
by every sublayer (compute width, input and output dimension of $F_i$) share the same width $d$.

\subsection{Decoupling Residual Width and Compute Width}
To decouple residual width and compute width, we introduce a wider residual stream
$h_i\in\mathbb{R}^{D}$ with $D\ge d$ and $n \mathrel{:=} \frac{D}{d}$ denotes the residual expansion ratio.
Since every $F_i$ keeps operating at
compute width $d$, there needs a read map
$R_i:\mathbb{R}^{D}\rightarrow\mathbb{R}^d$ to project the carried state into the
working space, and a write map
$W_i:\mathbb{R}^d\rightarrow\mathbb{R}^D$ to return the sublayer output to the
residual space. So the vanilla residual in Equation~\eqref{eq:vanilla-residual} becomes
\begin{equation}
  x_i=R_ih_{i-1},
  \qquad
  h_i
  =
  h_{i-1}+W_iF_i\!\left(\operatorname{RMSNorm}_i(x_i)\right),
  \qquad i=1,\ldots,2L .
  \label{eq:hires-compact}
\end{equation}
Setting $D=d$ and $R_i=W_i=I$ recovers the vanilla recurrence.

Residual-expansion methods typically introduce lightweight interfaces by sharing mixing weights across streams or coordinate groups. HiRes instead adopts
the simplest choice: dense projections $R_i\in\mathbb{R}^{d\times D}$ and $W_i\in\mathbb{R}^{D\times d}$. At fixed $D$ and $d$ these are the most general linear
interfaces, imposing no structural inductive biases like weight sharing, providing full connectivity between the expanded residual state and the working space.
Empirically, HiRes achieves competitive performance at a smaller expansion ratio than methods with structured interfaces, while incurring lower activation memory and I/O costs (Section~\ref{sec:priorart-comparison}).

\subsection{Hierarchical Residual Recurrence}

Applying dense interfaces at every sublayer would add
$\Theta(LDd)$ parameters and FLOPs, which is comparable to the
$\Theta(Ld^2)$ cost of the Transformer layers themselves and thus unacceptable.
HiRes therefore invokes them only at block boundaries, yielding a hierarchical structure.

HiRes partitions the $L$ layers into $B$ consecutive blocks
$\mathcal{B}_1,\ldots,\mathcal{B}_B$. Block $b$ contains $\ell_b$ layers,
where $\sum_{b=1}^{B}\ell_b=L$. For simplicity, we use equal-sized contiguous blocks with
$\ell_b=L/B$.

Inside each block $b$, HiRes keeps the inner PreNorm recurrence unchanged as the vanilla residual in Equation~\eqref{eq:vanilla-residual}:
\begin{equation}
  x_{b,i}
  =
  x_{b,i-1}
  +
  F_{b,i}\!\left(
    \operatorname{RMSNorm}_{b,i}(x_{b,i-1})
  \right),
  \qquad i=1,\ldots,2\ell_b .
  \label{eq:inner-residual}
\end{equation}
We write $\Phi_b$ for the composition of these recurrences, so that
$x_{b,2\ell_b}=\Phi_b(x_{b,0})$.

Between blocks, HiRes introduces an outer residual and normalizes it in PostNorm style,
transforming the layerwise recurrence in Equation~\eqref{eq:hires-compact} into a blockwise one:
\begin{equation}
  x_{b,0}=R_bh_{b-1},
  \qquad
  h_b = \operatorname{RMSNorm}_b(h_{b-1} + W_b\Phi_b(x_{b,0})),
  \qquad b=1,\ldots,B .
  \label{eq:outer-residual}
\end{equation}

Equations~\eqref{eq:inner-residual} and~\eqref{eq:outer-residual} define HiRes' two-level residual hierarchy:
a $d$-dimensional PreNorm inner residual and a $D$-dimensional PostNorm outer residual, illustrated in Figure~\ref{fig:hires-overview}.
Grouping layers into blocks amortizes the dense-interface overhead across depth, reducing it from $\Theta(LDd)$ to $\Theta(BDd)$.
Layers within each block thus share the same boundary interfaces and 
$d$-dimensional coordinate system: the block pulls a subspace of the outer residual state, transforms it through several layers, and commits the result back.

To connect the embedding and the language modeling head to the outer residual state, we use an embedding write projection
$E\in\mathbb{R}^{D\times d}$ and a head read projection $H\in\mathbb{R}^{d\times D}$,
\begin{equation}
  h_0=E\,\operatorname{Embed}(t),
  \qquad
  x_{\mathrm{final}}=H\,h_B,
  \label{eq:endpoint-maps}
\end{equation}
where $t$ is the input token.
Appendix~\ref{app:resource-accounting} details the exact parameter, FLOP,
memory, and I/O overheads, and Appendix Algorithm~\ref{alg:hires-forward}
provides PyTorch-style pseudocode for the forward pass.

\subsection{Design Choices}
\label{sec:design-choices}
\paragraph{Normalization at block boundaries.}
Boundary normalization provides residual regulation in HiRes: without it, the
outer residual stream would be unbounded as it accumulates block outputs.
Keeping PreNorm within blocks and applying PostNorm between them combines
PreNorm's optimization stability during the backward pass~\citep{wang2019learning,xiong2020layer}
with PostNorm's periodic control of the outer residual scale during the forward pass~\citep{kim2025periln}.

Setting \(D=d\) and \(R_b=W_b=I\) in Equation~\eqref{eq:outer-residual} and
removing the outer residual accumulation yields hierarchical normalization (HiNorm):
\begin{equation}
  x_{b+1,0} = \operatorname{RMSNorm}_b(\Phi_b(x_{b,0})),
  \qquad b=1,\ldots,B ,
  \label{eq:hierarchical-normalization}
\end{equation}
which inserts one RMSNorm into the residual stream every \(L/B\) layers. HiNorm
isolates regulation from residual expansion and serves as a control in
our experiments.

HiNorm alone beats the baseline and reaches minimum loss at an
intermediate \(B=4\) (Appendix Figure~\ref{fig:hinorm-frequency}).
In Appendix Figure~\ref{fig:activation-norm}, the residual scale grows monotonically 
in the vanilla baseline and in HiRes without boundary normalization. 
HC, VWN, and
AttnRes~\citep{zhu2025hyperconnections,seed2025vwn,kimiteam2026attentionresiduals}
regulate the scale implicitly through residual mixing learned at every layer,
whereas HiRes regulates it explicitly with boundary normalizations:
the scale grows only within each block and is reset at every boundary.

\begin{figure}[!t]
  \centering
  \includegraphics[width=\textwidth]{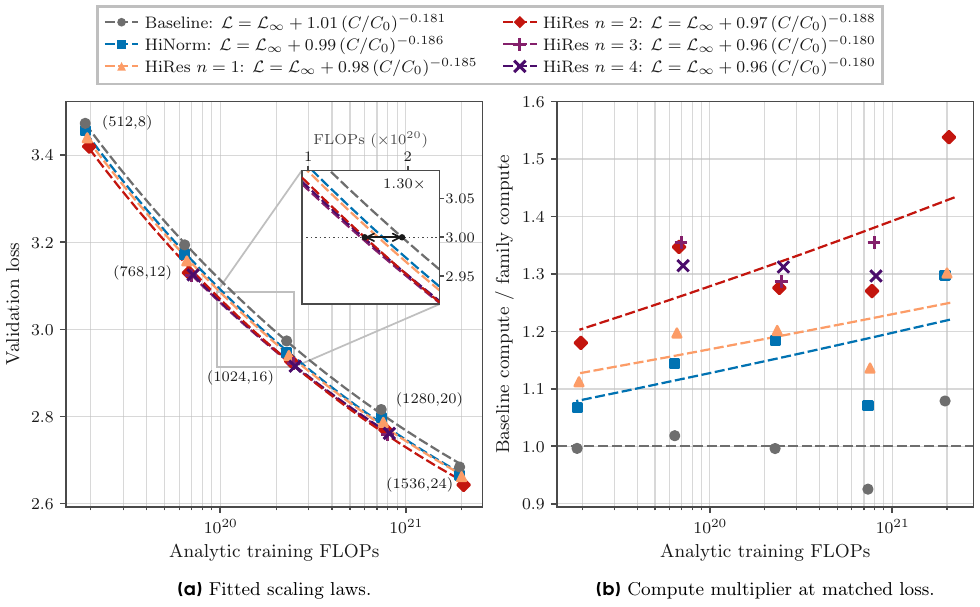}
  \caption{\textbf{HiRes achieves better scaling properties.}
  (a) Validation loss versus training FLOPs with fitted laws.
  At $\mathcal{L}=3.00$ (inset), the baseline needs $1.30\times$ the FLOPs of
  HiRes with $n=2$.
  (b) Compute multiplier at matched loss: baseline FLOPs over family
  FLOPs, at runs (markers) and along the fitted laws (lines).}
  \label{fig:scaling-compute}
\end{figure}

\paragraph{Interface stabilization.}
We empirically observe that randomly initialized interfaces degrade performance early in training (Appendix~\ref{sec:alpha}),
which we hypothesize arises from a seesaw dynamic: the interfaces learn each block's working coordinate system while the inner layers simultaneously learn to transform representations within that evolving space.
Motivated by prior identity-biased residual gates
~\citep{bachlechner2020rezero,touvron2021layerscale}, we stabilize the
interfaces by interpolating between identity and learned mappings using
learnable gating scalars:
\begin{equation}
  R_b = \bigl(1-\sigma(\alpha^{R}_b)\bigr)\Pi + \sigma(\alpha^{R}_b) \widetilde{R}_b,
  \qquad
  W_b = \bigl(1-\sigma(\alpha^{W}_b)\bigr)\Pi^{\top} + \sigma(\alpha^{W}_b) \widetilde{W}_b,
  \label{eq:interface-stabilization}
\end{equation}
where $\Pi=\bigl[I_{d\times d}, \mathbf{0}_{d\times (D-d)}\bigr]\in\mathbb{R}^{d\times D}$ truncates the outer state to its first $d$ coordinates, $\Pi^{\top}$ zero-pads back to width $D$, and $\sigma$ is the logistic sigmoid. Each interface has an
independent learned scalar logit initialized such that
$\sigma(\alpha)$ is close to $0$ (we set it to $0.1$ in the experiments).

This parameterization initializes the interfaces near identity, providing the inner layers with a relatively stable coordinate system early in training.
The interfaces can then gradually specialize and make increasing use of the expanded residual state parameterized by $\widetilde{R}_b$ and $\widetilde{W}_b$.

\section{Experiments}
\label{sec:experiments}
\subsection{Setup}
\label{sec:exp-setup}

We conduct the primary architecture study using the OLMoE
codebase~\citep{muennighoff2024olmoe}. All models use
RMSNorm~\citep{zhang2019rmsnorm}, RoPE~\citep{su2024roformer},
grouped-query attention (GQA)~\citep{ainslie2023gqa}, and a
SwiGLU feed-forward network~\citep{shazeer2020glu}. We train with
AdamW~\citep{loshchilov2019adamw} and a cosine learning-rate
schedule with warmup, use a $4096$-token context, and initialize
weights with standard deviation $0.02$.

The default HiRes configuration uses residual expansion ratio $n=D/d=2$ and block count
$B=4$. We ablate both choices in Section~\ref{sec:ablation}.

For architecture comparisons, we average training loss over the final $1000$ steps.
Validation loss and downstream accuracy are averaged over the final
$3$ evaluations, $1000$ steps apart, and across $11$ held-out domains
and $12$ tasks, respectively. The full experimental setup is detailed in
Appendix~\ref{app:experimental-details}.

\subsection{Scaling on Model Size}
\label{sec:scaling}

We evaluate model-size scaling over five model sizes, jointly
increasing inner width and depth from $(d,L)=(512,8)$ to $(1536,24)$ with the parameter count spanning from $70$M to $626$M.
The detailed scaling ladder can be found in Table~\ref{tab:scalingraw}.
We compare six fixed architecture families: the vanilla residual
baseline; HiRes with \(n=2\), \(n=3\), and \(n=4\); and two reduced forms of
HiRes, namely HiNorm, which keeps only the boundary
normalization, and equal-width HiRes with \(n=1\), which adds the interfaces
and the outer residual without widening. All HiRes variants use
\(B=4\).

Since HiRes introduces modest parameter overhead, we compare models by fitting a loss-compute scaling law
~\citep{kaplan2020scaling,hoffmann2022chinchilla} in analytic training FLOPs: $C = T(6P+12Lsd)$,
where $P$ is the total parameter count, $T$ is the number of training tokens,
$L$ is the number of layers, $s$ is the sequence length, and $d$ is the inner
width. We jointly fit the observations with the scaling law
$\mathcal{L}_f(C)=\mathcal{L}_{\infty}+A_f(C/C_0)^{-\alpha_f}$, with a shared irreducible loss and family-specific amplitude and exponent. Here $C_0 = 10^{20}$ FLOPs is the reference compute and $\mathcal{L}_{\infty}=2.101$ is the fitted irreducible loss.

Figure~\ref{fig:scaling-compute} separates the two sources of the HiRes gain: residual regulation and residual expansion.
HiNorm consistently shifts the vanilla residual scaling
curve downward. Widening the residual state ($n=2$) brings a further gain at every
scale. Equal-width HiRes ($n=1$) remains close to HiNorm, indicating that the
hierarchy interfaces alone add little benefit. A clearer improvement appears when
the outer state expands from $D=d$ to $D=2d$, while the $n=3$ and $n=4$
trajectories remain close to $n=2$, indicating diminishing returns. Regulation
and widening thus contribute complementary gains. HiRes with $n=2$ combines
both, and the fitted scaling laws estimate that the baseline requires
\(1.30\times\) the FLOPs of HiRes to reach a validation loss of \(3.00\).

\Needspace{8\baselineskip}
\subsection{Scaling on Depth}
\label{sec:depth-scaling}

\begin{wrapfigure}{r}{0.45\textwidth}
  \vspace{-\intextsep}
  \centering
  \includegraphics[width=\linewidth]{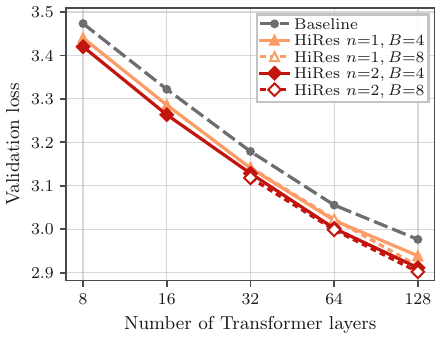}
  \caption{\textbf{HiRes scales better with depth.}
  HiRes wins at every depth and bends less at $L=128$; $B=8$ stays nearly straight.}
  \label{fig:scaling-depth}
\end{wrapfigure}

We next examine depth scaling properties of HiRes by fixing the inner width at $d=512$ and the training
budget at $30$B tokens while increasing the number of Transformer layers from
$8$ to $128$.

Figure~\ref{fig:scaling-depth} plots validation loss against depth
on a logarithmic axis, where ideal compute scaling should appear as a straight line.
HiRes not only consistently outperforms the vanilla residual baseline at every depth, but also maintains better scaling properties.
The baseline bends away from its scaling curve at the last doubling from $L=64$
to $L=128$.
With $B=4$, HiRes gives up less of its per-doubling reduction than the
baseline, especially at $n=2$, indicating that regulating residual accumulation and widening the residual state can hold more layers.
Raising the block count to $B=8$ further preserves the scaling close to
a straight line through $L=128$, suggesting more blocks for deeper models.

\WFclear
\Needspace{8\baselineskip}
\subsection{Comparison with Other Residual Expansion Methods}
\label{sec:priorart-comparison}

\begin{table}[!b]
  \caption{\textbf{HiRes versus other residual-expansion methods at
  $(d,L)=(768,12)$ and 50B tokens.}
  Suffixes give the expansion rate for HC/mHC/VWN and the block count for
  Block AttnRes. With a $2\times$ wider residual state,
  HiRes is competitive even with the $4\times$ variants of HC, mHC, and VWN. \textbf{Bold}/\underline{underlined}: best/second-best among
  residual-expansion methods. Appendix~\ref{app:priorart} details the
  configurations and lists their parameters and training FLOPs.}
  \label{tab:priorart-broad}
  \centering
  \small
  \setlength{\tabcolsep}{3pt}
  \begin{tabular}{@{}l*{10}{c}@{}}
    \toprule
    Metric & Baseline & HC-2 & HC-4 & mHC-2 & mHC-4 & VWN-2 & VWN-4 &
      \twolinehead{Full}{AttnRes} & \twolinehead{Block}{AttnRes-4} & HiRes \\
    \midrule
    Train loss & 3.2228 & 3.1773 & 3.1870 & 3.1941 & 3.1625 &
      \underline{3.1584} & \textbf{3.1578} & 3.1938 & 3.1943 & 3.1645 \\
    Val.\ loss & 3.1941 & 3.1649 & 3.1533 & 3.1871 & 3.1459 &
      3.1400 & \textbf{3.1228} & 3.1566 & 3.1649 & \underline{3.1300} \\
    Avg.\ Acc.\ (\%) & 46.91 & \underline{47.25} & 46.26 & 46.66 & \textbf{47.33} &
      47.03 & 45.86 & 46.17 & 46.13 & 47.15 \\
    \bottomrule
  \end{tabular}
\end{table}

\begin{table}[!t]
  \caption{\textbf{HiRes stays competitive across scales.}
  Activation I/O is the analytic count per token per sublayer for optimized
  implementations (Appendix~\ref{app:resource-accounting}); peak memory and throughput are
  measured with each method's optimized Triton implementation
  (Appendix~\ref{app:systems}), whereas HiRes is nearly as fast with vanilla PyTorch implementation
  (Appendix Table~\ref{tab:sys-stage}). \textbf{Bold}/\underline{underlined}:
  best/second-best among residual-expansion methods.}
  \label{tab:priorarttransfer}
  \centering
  \small
  \setlength{\tabcolsep}{5pt}
  \renewcommand{\arraystretch}{0.94}
  \begin{tabular}{llccccc}
    \toprule
    Backbone & Metric & Baseline & mHC-4 & VWN-4 & Block AttnRes-4 & HiRes \\
    \midrule
    \multirow{6}{*}{$(768,12)$}
      & Train loss & 3.2228 & \underline{3.1625} & \textbf{3.1578} & 3.1943 & 3.1645 \\
      & Val.\ loss & 3.1941 & 3.1459 & \textbf{3.1228} & 3.1649 & \underline{3.1300} \\
      & Avg.\ Acc.\ (\%) & 46.91 & \textbf{47.33} & 45.86 & 46.13 & \underline{47.15} \\
      & Activation I/O & 3$d$ & 17$d$ & 17$d$ & \underline{5.25$d$} & \textbf{4.58$\boldsymbol{d}$} \\
      & Peak memory (GiB) & 16.31 & 19.71 & 20.89 & \underline{17.18} & \textbf{16.86} \\
      & Throughput (M tok/s) & 0.682 & 0.590 & 0.556 & \textbf{0.639} & \underline{0.625} \\
    \midrule
    \multirow{6}{*}{$(1024,16)$}
      & Train loss & 3.0108 & \textbf{2.9633} & 2.9678 & 2.9912 & \underline{2.9663} \\
      & Val.\ loss & 2.9734 & \textbf{2.9205} & \underline{2.9231} & 2.9547 & 2.9271 \\
      & Avg.\ Acc.\ (\%) & 48.76 & 49.16 & \textbf{49.57} & \underline{49.45} & 49.15 \\
      & Activation I/O & 3$d$ & 17$d$ & 17$d$ & \underline{5.19$d$} & \textbf{4.19$\boldsymbol{d}$} \\
      & Peak memory (GiB) & 21.49 & 27.28 & 28.84 & \underline{22.96} & \textbf{22.22} \\
      & Throughput (M tok/s) & 0.419 & 0.370 & 0.354 & \textbf{0.398} & \underline{0.389} \\
    \midrule
    \multirow{6}{*}{$(1280,20)$}
      & Train loss & 2.8524 & 2.8157 & \textbf{2.8068} & 2.8366 & \underline{2.8097} \\
      & Val.\ loss & 2.8158 & \underline{2.7671} & \textbf{2.7533} & 2.7871 & 2.7697 \\
      & Avg.\ Acc.\ (\%) & 51.90 & \underline{52.49} & 52.30 & 51.48 & \textbf{52.76} \\
      & Activation I/O & 3$d$ & 17$d$ & 17$d$ & \underline{5.15$d$} & \textbf{3.95$\boldsymbol{d}$} \\
      & Peak memory (GiB) & 28.07 & 36.89 & 38.85 & \underline{30.33} & \textbf{28.99} \\
      & Throughput (M tok/s) & 0.279 & 0.240 & 0.224 & \textbf{0.263} & \underline{0.259} \\
    \bottomrule
  \end{tabular}
  \vspace{-8pt}
\end{table}

We compare HiRes with other residual expansion methods with the recipe of $(d,L)=(768,12)$ for 50B-token training, namely
Hyper-Connections (HC)~\citep{zhu2025hyperconnections},
mHC~\citep{xie2025mhc}, Virtual Width Networks (VWN)~\citep{seed2025vwn}, and Attention
Residuals~\citep{kimiteam2026attentionresiduals}. Specifically, we test $2\times$ and $4\times$ wider residual states for HC/mHC/VWN,
full and $4$-block AttnRes, and $n=2,B=4$ HiRes.
We then test mHC-4, VWN-4, Block AttnRes-4,
and HiRes configurations on larger backbones. Appendix~\ref{app:priorart} gives the detailed implementations and complete results.

At $(d,L)=(768,12)$, all residual-expansion methods improve the vanilla
residual baseline in loss, and HiRes achieves comparable
performance to mHC, VWN, and AttnRes at a lower residual expansion rate,
as shown in Table~\ref{tab:priorart-broad}.

Table~\ref{tab:priorarttransfer} compares the vanilla residual baseline,
mHC, VWN, Block AttnRes, and HiRes across three backbone scales: $(d, L) = (768, 12), (1024, 16), (1280, 20)$. HiRes
remains within the leading quality range of these recent residual-expansion
methods. The same table reports the analytic I/O at each scale
(Appendix~\ref{app:resource-accounting}), together with throughput and memory from a unified
measurement of each method's highly optimized Triton implementation (Appendix~\ref{app:systems}).
HiRes has the lowest activation I/O among residual-expansion methods and closest
to the baseline's $3d$. Despite additional parameters, HiRes retains $91.7$--$93.1\%$ of baseline throughput at
$3.3$--$3.4\%$ memory overhead, which is on par with optimized AttnRes and more efficient than mHC/VWN. Moreover, HiRes' standard PyTorch implementation already reaches near-final performance,
while other methods rely on mechanism-specific kernel fusion.
Appendix~\ref{app:systems} reports the agentic optimization campaign and the latency
decomposition behind these numbers.

\begin{figure}[!b]
  \vspace{-6pt}
  \centering
  \includegraphics[width=\textwidth]{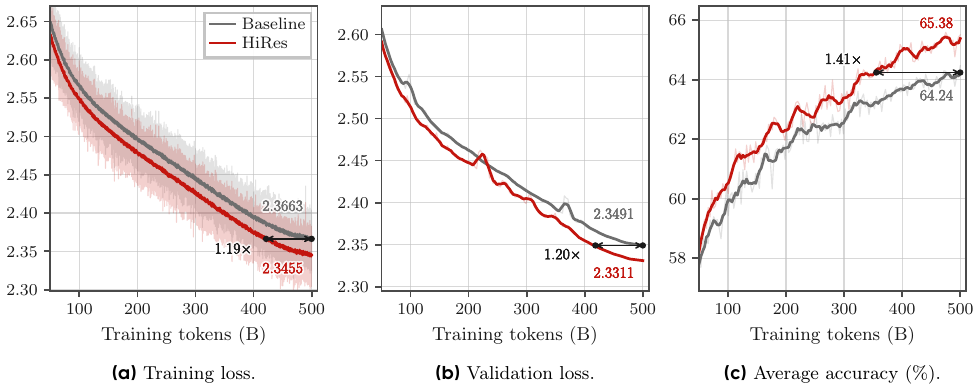}
  \caption{\textbf{HiRes on 500B-token OLMoE-1B-7B pretraining.}
  HiRes ends with $0.0208$ lower training loss, $0.0180$ lower validation loss,
  and a $1.14$-point higher average accuracy. See Appendix~\ref{app:long-horizon-moe} for details.}
  \label{fig:main-moe-dynamics}
  \vspace{-8pt}
\end{figure}

\subsection{Long-Horizon MoE Pretraining}
\label{sec:long-horizon-moe-pretraining}

To scale up, we evaluate HiRes on the OLMoE-1B-7B pretraining
recipe~\citep{muennighoff2024olmoe}, a sparse Mixture-of-Experts
model~\citep{shazeer2017moe,fedus2022switch}. The model contains
$6.9$B total parameters and $1.3$B active parameters per
token, and is trained for $500$B tokens. We compare the vanilla residual
baseline with HiRes using the $n=2$, $B=4$ configuration, which adds 83.9M
parameters ($6.5\%$ of the active parameters and $1.2\%$ of the total parameters); Appendix
Table~\ref{tab:moe-model-card} lists the full configuration.

As shown in Figure~\ref{fig:main-moe-dynamics}, HiRes achieves $0.0208$ lower training loss and $0.0180$ lower
validation loss than the vanilla residual baseline. It also performs better
on most downstream tasks and improves the downstream average accuracy by $1.14$ points.
These results establish that the HiRes design transfers to long-horizon sparse-MoE pretraining.

\section{Ablation Studies}

\subsection{Ablation on Block Count and Expansion Ratio}
\label{sec:ablation}

\begin{figure}[!t]
  \centering
  \begin{subfigure}[t]{0.49\textwidth}
    \centering
    \includegraphics[width=\linewidth]{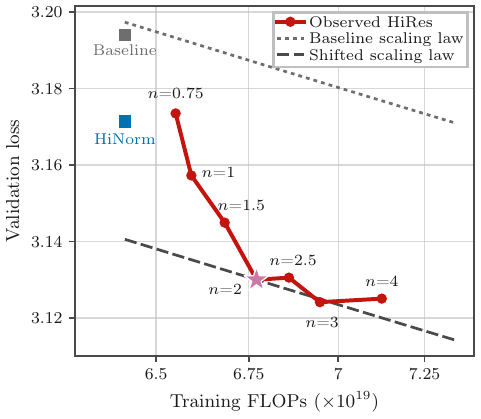}
    \caption{Expansion ratio $n$ at $B=4$.}
    \label{fig:ablation-expansion}
  \end{subfigure}
  \hfill
  \begin{subfigure}[t]{0.49\textwidth}
    \centering
    \includegraphics[width=\linewidth]{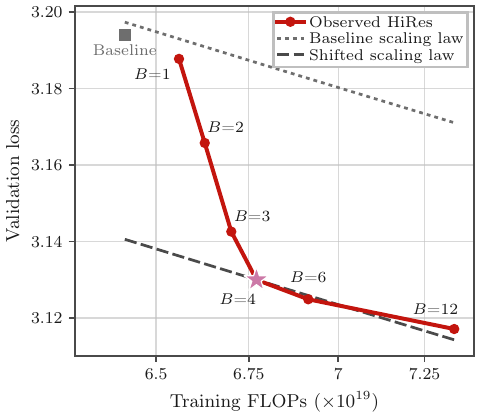}
    \caption{Block count $B$ at $n=2$.}
    \label{fig:ablation-blocks}
  \end{subfigure}
  \caption{\textbf{Ablation on expansion ratio $n$ and block count $B$ in HiRes.}
  Beyond the default $n=2$, $B=4$ (star),
  gains roughly follow the shifted scaling law (dashed), which translates the
  baseline scaling law (dotted) through the star, accounting for the benefit from increasing parameters.}
  \label{fig:nbablation}
\end{figure}

We ablate the two structural hyperparameters of HiRes: the residual expansion
ratio $n$ and the number of blocks $B$. All runs use $(d,L)=(768,12)$ and
50B training tokens. The expansion sweep fixes $B=4$ and varies $n=D/d$ from
$0.75$ to $4$, while the block sweep fixes $n=2$ and varies $B$ from $1$ to
$12$. Figure~\ref{fig:nbablation} reports both sweeps.

\paragraph{Expansion ratio.}
In Figure~\ref{fig:ablation-expansion}, validation loss rapidly decreases as the
expansion ratio increases to $n=2$, after which the slope slows. The wider $n=3$ and $n=4$ settings provide
marginal gains, largely following the shifted scaling law, which accounts for the benefit from extra parameters. 
Thus $n=2$ is the empirical knee of the quality-overhead curve:
it captures most of the benefit of residual expansion while keeping the
interface cost moderate.
On the other side, at $n=0.75$ the
outer state is narrower than the layer compute width, and it is the only point
in the sweep that fails to improve over HiNorm, even with more parameters.
This result identifies residual width,
rather than the interfaces themselves, as the main source of the gain.

\paragraph{Number of blocks.}
In Figure~\ref{fig:ablation-blocks}, increasing $B$ also improves validation
loss, with the largest gains occurring as the hierarchy is refined from one to
four blocks. Beyond $B=4$, more
frequent outer-state updates show diminishing returns similar to those of the
expansion ratio.
We therefore retain $B=4$ as the empirical knee.

\subsection{Architecture Ablations}
\label{sec:architecture-ablation}

We subtract one component at a time from the complete
$n=2,B=4$ HiRes model, testing boundary normalization, outer residual,
and interface learning against the same reference architecture, as shown in Table~\ref{tab:architecture-ablation}; HiNorm and equal-width HiRes
($n=1$) serve as reduced references. All ablations
use $(d,L)=(768,12)$ and 50B training tokens.

The ablations support the following conclusions.
(i) The interfaces must be learnable. Freezing the read/write projections
raises validation loss by \(0.031\), so widening residual with fixed maps does not work. (ii) Widening helps even without normalization. Removing boundary RMSNorm raises
validation loss by \(0.025\), yet this

\begin{wraptable}{r}{0.5\textwidth}
  \vspace{-\intextsep}
  \caption{\textbf{Ablations on HiRes components.}
  Removing or freezing any one component of HiRes ($n=2$, $B=4$) hurts.}
  \label{tab:architecture-ablation}
  \centering
  \small
  \setlength{\tabcolsep}{3pt}
  \renewcommand{\arraystretch}{0.96}
  \begin{tabular}{lcc}
    \toprule
    Variant & Train loss & Val.\ loss \\
    \midrule
    Baseline & 3.2228 & 3.1941 \\
    HiNorm & 3.1956 & 3.1714 \\
    HiRes $n=1,B=4$ & 3.1869 & 3.1573 \\
    \midrule
    HiRes $n=2,B=4$ & 3.1645 & 3.1300 \\
    \quad-- Frozen random $R/W$ & 3.1905 & 3.1614 \\
    \quad-- w/o boundary RMSNorm & 3.1874 & 3.1552 \\
    \quad-- w/o outer residual & 3.1899 & 3.1616 \\
    \quad-- w/o interface stabilization & 3.1686 & 3.1375 \\
    \bottomrule
  \end{tabular}
\end{wraptable}
widening variant still improves on the baseline by
\(0.039\).
(iii) Outer residual is necessary. Removing the outer residual raises validation loss by
\(0.032\), because each write-read pair between adjacent blocks then collapses
into a single $d\rightarrow D \rightarrow d$ MLP with boundary RMSNorm as its
activation, leaving the residual state effectively unexpanded.
(iv) Interface stabilization gives a modest final gain. Disabling it raises
validation loss by \(0.008\), a relatively modest gap. Nevertheless, its advantage is much larger early in
training (Appendix~\ref{sec:alpha}), which is the main reason we adopt it.

\begingroup
\titlespacing*{\section}{0pt}{3.5ex plus 1ex minus .2ex}{1ex plus .2ex}
\titlespacing*{\paragraph}{0pt}{1ex plus .2ex minus .2ex}{1em}
\section{Related Work}
\label{sec:related}

\paragraph{Residual-state expansion.}
The closest methods expand the residual state, but typically by replacing
vanilla addition with a learned, per-layer operator. AltUp widens the
representation into sub-blocks and updates one sub-block at backbone width while a
lightweight predictor updates the rest~\citep{baykal2023altup}.
Hyper-Connections (HC) maintains multiple residual streams with learned
layer-dependent read, write, and residual mixing
weights~\citep{zhu2025hyperconnections}; Virtual Width Networks (VWN) generalizes HC by
constructing fine-grained fractional streams, enabling more expressive mixing~\citep{zhu2025frac,seed2025vwn};
mHC constrains the residual stream transition to a doubly
stochastic manifold for stability~\citep{xie2025mhc}. Attention Residuals instead learns
content-dependent attention over earlier layer or block
representations~\citep{kimiteam2026attentionresiduals}. Relative to this
family, HiRes is simpler in mechanism: one wide outer state, dense read/write
only at block boundaries, and additive residual transport. It maintains more expressive interfaces between the expanded residual
and the fixed-width inner state, and amortizes the cost across depth.

\paragraph{Normalization in Transformers.}
The original Transformer normalizes after each residual addition (PostNorm),
whereas modern language models typically use PreNorm to keep an uninterrupted
additive path through depth
~\citep{vaswani2017attention,ba2016layernorm,nguyen2019transformers,xiong2020layer,zhang2019rmsnorm}.
There exists a well-known trade-off between PreNorm and PostNorm:
PostNorm controls activation scale but can attenuate gradients,
while PreNorm eases optimization but lets the residual state grow
unbounded~\citep{wang2022deepnet}.
Hybrid placements pursue both, adding normalization inside the
residual branches~\citep{shleifer2021normformer}, around each
sublayer~\citep{kim2025periln}, mixed within a
sublayer~\citep{zhuo2025hybridnorm}, across depth~\citep{li2025mixln}, or
across coupled PreNorm-like and PostNorm-like
streams~\citep{li2026siamesenorm}. HiRes differs in granularity: complete PreNorm
layers run inside a block, and PostNorm is applied only after the block-level
outer residual addition, preserving ordinary inner updates while periodically
resetting the outer-state scale.

\paragraph{Cross-layer connectivity and state aggregation.}
Prior work enriches information flow across depth through gated shortcuts,
identity paths, or dense connectivity
~\citep{srivastava2015highway,he2016deep,huang2017densely}. Transformer
variants further aggregate earlier representations using static weights,
element-wise gates, reused values, or low-rank projections
~\citep{pagliardini2024denseformer,fang2023mrla,zhou2025valueresidual,menghani2025laurel},
while MUDDFormer mixes decoupled streams and Deep Delta Learning maintains a
matrix-valued state~\citep{xiao2025muddformer,zhang2026ddl}. HiRes instead
uses hierarchical connectivity, combining a persistent wide outer state
across blocks with narrow inner states within each block.
\par\endgroup

\section{Conclusion}
We introduced Hierarchical Residuals (HiRes), a deliberately minimal
architecture that decouples residual-state width from per-layer compute
width. HiRes carries a wide persistent state across layer blocks, performs
ordinary Transformer computation in a narrower inner space, and connects the
two through learned read/write projections at block boundaries. This
hierarchical design uses only dense linear maps, normalization, and residual
addition, requiring neither specialized kernels nor changes to the inner
Transformer layers.

Across model sizes and depths, HiRes consistently
outperforms the vanilla residual baseline and is competitive with other residual expansion methods including HC, VWN, mHC, and AttnRes.
These results demonstrate that residual-state width need not be tied to per-layer
compute width and that their conventional equality can be suboptimal.
HiRes therefore establishes residual width as a practical
scaling variable that can be exploited through a simple yet effective hierarchical design.

\clearpage
\beginappendix
\section{Implementation Pseudocode}
\label{app:implementation}

HiRes can be implemented as a simple extension of a vanilla Transformer
by changing fewer than ten lines of code in the model's \texttt{forward} method,
as illustrated in Algorithm~\ref{alg:hires-forward}.

\begin{algorithm}[h]
  \caption{\textbf{PyTorch-style pseudocode of the HiRes forward pass.}}
  \label{alg:hires-forward}
\begin{lstlisting}[style=pytorch]
def forward(self, input_ids):
    x = self.embed(input_ids)
    x_outer = self.embed_proj(x)  # up-project d -> D
    for b in range(self.num_blocks):
        # Read at block entry, down-project D -> d
        x_inner = self.read_proj[b](x_outer)
        # Standard PreNorm inner residual
        for i in range(self.num_layers // self.num_blocks):
            x_inner = x_inner + self.attn[b][i](self.attn_norm[b][i](x_inner))
            x_inner = x_inner + self.ffn[b][i](self.ffn_norm[b][i](x_inner))
        # Write at block exit, up-project d -> D
        x_outer = self.outer_norm[b](x_outer + self.write_proj[b](x_inner))
    x = self.lm_head_proj(x_outer)  # down-project D -> d
    logits = self.lm_head(self.final_layernorm(x))
    return logits
\end{lstlisting}
\end{algorithm}

\section{Experimental Setup}
\label{app:experimental-details}

All dense experiments use decoder-only Transformers implemented in the OLMoE
codebase~\citep{muennighoff2024olmoe}.
Models use weight-tied
input/output embeddings, RMSNorm~\citep{zhang2019rmsnorm},
RoPE~\citep{su2024roformer}, SwiGLU~\citep{shazeer2020glu} with preactivation
width
$d_{\mathrm{ff}}=4d$, grouped-query attention~\citep{ainslie2023gqa} with head
dimension 64 and 4
KV heads, and per-head QK-norm~\citep{dehghani2023vit22b}. Training uses
AdamW~\citep{loshchilov2019adamw} with $\beta_1=0.9$, $\beta_2=0.95$,
$\epsilon=10^{-8}$, weight decay
$0.1$, and gradient-norm clipping at $1.0$. The learning rate uses cosine
decay to $0.1$ of the peak rate after a linear warmup over 2\% of training tokens. Models use
a 4,096-token sequence length and initialization standard deviation $0.02$.
Table~\ref{tab:training-recipes} lists the scale-dependent settings.

The training corpus is the OLMo2 stage-1
mixture, tokenized with the Dolma2
tokenizer and its 100,352-token vocabulary~\citep{olmo2025olmo2}. All compared architectures at a
given scale reuse the same data order and global sample-index array.

Validation loss is averaged over 11 held-out domains from
Paloma~\citep{magnusson2024paloma}: C4~\citep{raffel2020t5},
Dolma Books, Dolma Common Crawl, Dolma Pes2o, Dolma Reddit, Dolma Stack,
Dolma Wiki~\citep{soldaini2024dolma}, ICE, M2D2
S2ORC~\citep{reid2022m2d2,lo2020s2orc}, Pile~\citep{gao2020pile}, and
WikiText-103~\citep{merity2017pointer}. Downstream accuracy is averaged over 12 tasks: MMLU~\citep{hendrycks2021mmlu}, ARC-Challenge and
ARC-Easy~\citep{clark2018arc}, BoolQ~\citep{clark2019boolq},
CommonsenseQA~\citep{talmor2019commonsenseqa}, COPA~\citep{roemmele2011copa},
HellaSwag~\citep{zellers2019hellaswag},
OpenBookQA~\citep{mihaylov2018openbookqa}, PIQA~\citep{bisk2020piqa},
SciQ~\citep{welbl2017sciq}, SocialIQA~\citep{sap2019socialiqa}, and
WinoGrande~\citep{sakaguchi2021winogrande}.

\begin{table}[!htbp]
  \caption{\textbf{Scale-dependent training recipes.}}
  \label{tab:training-recipes}
  \centering
  \small
  \setlength{\tabcolsep}{7pt}
  \begin{tabular}{lrrrrr}
    \toprule
    Configuration $(d,L)$
      & $(512,8)$ & $(768,12)$ & $(1024,16)$ & $(1280,20)$ & $(1536,24)$ \\
    \midrule
    Attention heads & 8 & 12 & 16 & 20 & 24 \\
    KV heads & 4 & 4 & 4 & 4 & 4 \\
    FFN width & 2,048 & 3,072 & 4,096 & 5,120 & 6,144 \\
    Baseline params & 70.264M & 138.433M & 245.403M & 403.757M & 626.078M \\
    Global batch size & 128 & 128 & 256 & 384 & 512 \\
    Peak learning rate
      & $8.0{\times}10^{-4}$ & $6.5{\times}10^{-4}$
      & $5.7{\times}10^{-4}$ & $5.1{\times}10^{-4}$
      & $4.6{\times}10^{-4}$ \\
    Training tokens & 30B & 50B & 100B & 200B & 350B \\
    \bottomrule
  \end{tabular}
\end{table}

\FloatBarrier
\section{Detailed Experimental Results}
\label{app:additional-scaling}

\subsection{Scaling on Model Size}
Following the model-size study in Section~\ref{sec:scaling}, Table~\ref{tab:scalingraw} provides the complete results at every evaluated scale.

\begin{table}[!htbp]
  \caption{\textbf{Parameters, training FLOPs, and losses for model size scaling.}
  At every model size, loss falls from the baseline to HiNorm,HiRes $n=1$, and HiRes $n=2$. HiNorm and HiRes use $B=4$.}
  \label{tab:scalingraw}
  \centering
  \footnotesize
  \setlength{\tabcolsep}{3pt}
  \begin{tabular}{lllrrrrrr}
    \toprule
    & & & & & \multicolumn{4}{c}{HiRes} \\
    \cmidrule(lr){6-9}
    $(d,L)$ & Tokens & Metric & Baseline & HiNorm & $n=1$ & $n=2$ & $n=3$ & $n=4$ \\
    \midrule
    \multirow{4}{*}{$(512,8)$} & \multirow{4}{*}{30B}
      & Params & 70.264M & 70.266M & 72.888M & 75.511M & -- & -- \\
      & & FLOPs ($10^{19}$) & 1.8691 & 1.8691 & 1.9163 & 1.9635 & -- & -- \\
      & & Train loss & 3.4768 & 3.4617 & 3.4515 & 3.4314 & -- & -- \\
      & & Val.\ loss & 3.4733 & 3.4564 & 3.4401 & 3.4201 & -- & -- \\
    \midrule
    \multirow{4}{*}{$(768,12)$} & \multirow{4}{*}{50B}
      & Params & 138.433M & 138.436M & 144.334M & 150.235M & 156.136M & 162.037M \\
      & & FLOPs ($10^{19}$) & 6.4186 & 6.4187 & 6.5957 & 6.7727 & 6.9497 & 7.1267 \\
      & & Train loss & 3.2228 & 3.1956 & 3.1869 & 3.1645 & 3.1586 & 3.1591 \\
      & & Val.\ loss & 3.1941 & 3.1714 & 3.1573 & 3.1300 & 3.1242 & 3.1251 \\
    \midrule
    \multirow{4}{*}{$(1024,16)$} & \multirow{4}{*}{100B}
      & Params & 245.403M & 245.407M & 255.892M & 266.382M & 276.872M & 287.362M \\
      & & FLOPs ($10^{19}$) & 22.780 & 22.780 & 23.409 & 24.039 & 24.669 & 25.299 \\
      & & Train loss & 3.0108 & 2.9827 & 2.9758 & 2.9663 & 2.9589 & 2.9539 \\
      & & Val.\ loss & 2.9734 & 2.9466 & 2.9401 & 2.9271 & 2.9219 & 2.9153 \\
    \midrule
    \multirow{4}{*}{$(1280,20)$} & \multirow{4}{*}{200B}
      & Params & 403.757M & 403.762M & 420.146M & 436.535M & 452.924M & 469.313M \\
      & & FLOPs ($10^{19}$) & 73.623 & 73.623 & 75.590 & 77.556 & 79.522 & 81.488 \\
      & & Train loss & 2.8524 & 2.8338 & 2.8245 & 2.8097 & 2.8015 & 2.8065 \\
      & & Val.\ loss & 2.8158 & 2.7972 & 2.7865 & 2.7697 & 2.7590 & 2.7612 \\
    \midrule
    \multirow{4}{*}{$(1536,24)$} & \multirow{4}{*}{350B}
      & Params & 626.078M & 626.084M & 649.677M & 673.276M & -- & -- \\
      & & FLOPs ($10^{19}$) & 194.91 & 194.91 & 199.86 & 204.82 & -- & -- \\
      & & Train loss & 2.7295 & 2.7084 & 2.7019 & 2.6907 & -- & -- \\
      & & Val.\ loss & 2.6838 & 2.6647 & 2.6618 & 2.6427 & -- & -- \\
    \bottomrule
  \end{tabular}
\end{table}

\FloatBarrier
\subsection{Scaling on Depth}
In Section~\ref{sec:depth-scaling} we study depth scaling at fixed width;
Table~\ref{tab:scalingdepth} reports the corresponding results and the
follow-up sweep over the number of HiRes blocks.

\begin{table}[!htbp]
  \caption{\textbf{HiRes beats baseline at every depth, and $B=8$ helps most
  at $L=128$.} All models use $d=512$ and 30B tokens.}
  \label{tab:scalingdepth}
  \label{tab:depthblocks}
  \centering
  \footnotesize
  \setlength{\tabcolsep}{3.5pt}
  \begin{tabular}{llrrrrr}
    \toprule
    & & & \multicolumn{4}{c}{HiRes} \\
    \cmidrule(lr){4-7}
    $(d,L)$ & Metric & Baseline
      & $n=1,B=4$ & $n=1,B=8$ & $n=2,B=4$ & $n=2,B=8$ \\
    \midrule
    \multirow{4}{*}{$(512,8)$}
      & Params & 70.264M & 72.888M & -- & 75.511M & -- \\
      & FLOPs ($10^{19}$) & 1.8691 & 1.9163 & -- & 1.9635 & -- \\
      & Train loss & 3.4768 & 3.4515 & -- & 3.4314 & -- \\
      & Val.\ loss & 3.4733 & 3.4401 & -- & 3.4201 & -- \\
    \midrule
    \multirow{4}{*}{$(512,16)$}
      & Params & 89.148M & 91.771M & -- & 94.395M & -- \\
      & FLOPs ($10^{19}$) & 2.8126 & 2.8598 & -- & 2.9071 & -- \\
      & Train loss & 3.3334 & 3.2934 & -- & 3.2774 & -- \\
      & Val.\ loss & 3.3221 & 3.2860 & -- & 3.2635 & -- \\
    \midrule
    \multirow{4}{*}{$(512,32)$}
      & Params & 126.915M & 129.539M & 131.638M & 132.162M & 136.360M \\
      & FLOPs ($10^{19}$) & 4.7004 & 4.7476 & 4.7854 & 4.7948 & 4.8704 \\
      & Train loss & 3.1949 & 3.1521 & 3.1551 & 3.1421 & 3.1364 \\
      & Val.\ loss & 3.1790 & 3.1407 & 3.1421 & 3.1285 & 3.1182 \\
    \midrule
    \multirow{4}{*}{$(512,64)$}
      & Params & 202.449M & 205.073M & 207.172M & 207.696M & 211.895M \\
      & FLOPs ($10^{19}$) & 8.4759 & 8.5232 & 8.5609 & 8.5704 & 8.6459 \\
      & Train loss & 3.0695 & 3.0323 & 3.0318 & 3.0185 & 3.0162 \\
      & Val.\ loss & 3.0555 & 3.0202 & 3.0232 & 3.0019 & 2.9986 \\
    \midrule
    \multirow{4}{*}{$(512,128)$}
      & Params & 353.518M & 356.142M & 358.241M & 358.765M & 362.963M \\
      & FLOPs ($10^{19}$) & 16.027 & 16.074 & 16.112 & 16.121 & 16.197 \\
      & Train loss & 2.9918 & 2.9536 & 2.9258 & 2.9232 & 2.9141 \\
      & Val.\ loss & 2.9765 & 2.9387 & 2.9154 & 2.9111 & 2.9017 \\
    \bottomrule
  \end{tabular}
\end{table}

\FloatBarrier
\subsection{Comparison with Other Residual Expansion Methods}
\label{app:priorart}

\paragraph{Configurations.}
We detail the configurations compared in
Section~\ref{sec:priorart-comparison}. HC-$m$ and mHC-$m$ maintain $m$
width-$d$ residual streams. VWN-$m$ has the same total virtual width $md$ but
uses finer-grained streams: $2m$ streams of width $d/2$. We evaluate
$m\in\{2,4\}$. Full AttnRes retrieves from all preceding sublayer outputs,
whereas Block AttnRes-4 partitions the $2L$ residual sublayers into $4$
equal blocks and retains one representation per completed block. HiRes uses
expansion ratio $n=D/d=2$ and block count $B=4$ throughout.
Table~\ref{tab:priorart-cost} lists the parameters and training FLOPs of each
configuration.

\begin{table}[!htbp]
  \caption{\textbf{Parameters and training FLOPs of the compared residual expansion
  methods.} FLOPs are reported in units of $10^{19}$.}
  \label{tab:priorart-cost}
  \centering
  \footnotesize
  \setlength{\tabcolsep}{5pt}
  \begin{tabular}{lrrrrrr}
    \toprule
    & \multicolumn{2}{c}{$(768,12)$} & \multicolumn{2}{c}{$(1024,16)$}
    & \multicolumn{2}{c}{$(1280,20)$} \\
    \cmidrule(lr){2-3}\cmidrule(lr){4-5}\cmidrule(lr){6-7}
    Method & Params & FLOPs & Params & FLOPs & Params & FLOPs \\
    \midrule
    Baseline & 138.4M & 6.419 & 245.4M & 22.78 & 403.8M & 73.62 \\
    HC-2 & 138.5M & 6.421 & -- & -- & -- & -- \\
    HC-4 & 138.6M & 6.423 & -- & -- & -- & -- \\
    mHC-2 & 138.8M & 6.429 & -- & -- & -- & -- \\
    mHC-4 & 140.3M & 6.474 & 248.7M & 22.98 & 408.9M & 74.24 \\
    VWN-2 & 140.9M & 6.492 & -- & -- & -- & -- \\
    VWN-4 & 143.3M & 6.564 & 254.0M & 23.30 & 417.2M & 75.24 \\
    Full AttnRes & 138.5M & 6.420 & -- & -- & -- & -- \\
    Block AttnRes-4 & 138.5M & 6.420 & 245.5M & 22.78 & 403.9M & 73.64 \\
    HiRes & 150.2M & 6.773 & 266.4M & 24.04 & 436.5M & 77.56 \\
    \bottomrule
  \end{tabular}
\end{table}

\paragraph{Per-task downstream accuracy.}
Table~\ref{tab:priorart-tasks} reports the complete downstream-task breakdown
for the four residual-expansion methods transferred across all three backbone
scales.

\begin{table}[!htb]
  \caption{\textbf{Downstream performance for different residual expansion methods.} Best and second-best values are \textbf{bold} and \underline{underlined}. HiRes remains competitive across model scales.}
  \label{tab:priorart-tasks}
  \centering
  \footnotesize
  \setlength{\tabcolsep}{5pt}
  \renewcommand{\arraystretch}{0.92}
  \begin{tabular}{llccccc}
    \toprule
    Backbone & Task & Baseline & mHC-4 & VWN-4 & Block AttnRes-4 & HiRes \\
    \midrule
    \multirow{13}{*}{$(768,12)$}
      & MMLU & 27.87 & 27.24 & \underline{27.28} & \textbf{27.40} & 26.59 \\
      & ARC-Challenge & 25.98 & \textbf{24.75} & 22.97 & \underline{23.63} & \textbf{24.75} \\
      & ARC-Easy & 50.94 & \underline{51.93} & 50.41 & \textbf{52.05} & 51.11 \\
      & BoolQ & 59.67 & \underline{55.84} & 51.27 & \textbf{56.06} & 55.18 \\
      & CommonsenseQA & 32.05 & \underline{34.07} & 31.59 & 31.61 & \textbf{34.34} \\
      & COPA & 67.00 & \underline{68.00} & 65.00 & 62.33 & \textbf{70.67} \\
      & HellaSwag & 33.42 & \underline{34.83} & \underline{34.83} & 34.25 & \textbf{35.26} \\
      & OpenBookQA & 30.60 & \underline{29.40} & 29.07 & \textbf{30.53} & 29.27 \\
      & PIQA & 64.62 & \underline{65.61} & \underline{65.61} & 65.52 & \textbf{65.96} \\
      & SciQ & 76.10 & \textbf{81.50} & \underline{78.83} & 77.30 & \underline{78.83} \\
      & SocialIQA & 42.73 & \underline{42.84} & 42.39 & 42.15 & \textbf{43.09} \\
      & WinoGrande & 51.91 & \textbf{51.93} & \underline{51.12} & 50.70 & 50.72 \\
      \cmidrule(lr){2-7}
      & Avg.\ Acc. & 46.91 & \textbf{47.33} & 45.86 & 46.13 & \underline{47.15} \\
    \midrule
    \multirow{13}{*}{$(1024,16)$}
      & MMLU & 28.20 & \underline{28.82} & \textbf{28.89} & 28.52 & 28.40 \\
      & ARC-Challenge & 26.09 & 25.31 & 24.86 & \underline{26.20} & \textbf{26.98} \\
      & ARC-Easy & 56.43 & 54.56 & \underline{56.43} & \textbf{56.96} & 55.73 \\
      & BoolQ & 57.28 & 58.28 & \underline{59.94} & \textbf{62.04} & 56.38 \\
      & CommonsenseQA & 35.54 & 33.58 & \underline{35.05} & 34.21 & \textbf{35.93} \\
      & COPA & 69.00 & 69.00 & \underline{69.67} & \textbf{70.67} & 66.00 \\
      & HellaSwag & 40.18 & 42.02 & \textbf{42.14} & 40.74 & \underline{42.03} \\
      & OpenBookQA & 30.60 & \underline{31.87} & \textbf{32.00} & 30.93 & \textbf{32.00} \\
      & PIQA & 65.92 & \underline{67.99} & 67.27 & 66.47 & \textbf{68.61} \\
      & SciQ & 82.03 & \textbf{82.73} & \underline{82.20} & 81.60 & 81.57 \\
      & SocialIQA & 43.06 & \underline{43.52} & 43.50 & 42.53 & \textbf{43.72} \\
      & WinoGrande & 50.80 & 52.20 & \textbf{52.93} & 52.49 & \underline{52.51} \\
      \cmidrule(lr){2-7}
      & Avg.\ Acc. & 48.76 & 49.16 & \textbf{49.57} & \underline{49.45} & 49.15 \\
    \midrule
    \multirow{13}{*}{$(1280,20)$}
      & MMLU & 30.49 & 30.75 & \underline{30.89} & 30.56 & \textbf{30.95} \\
      & ARC-Challenge & 29.32 & \textbf{29.10} & \underline{28.65} & \underline{28.65} & 27.65 \\
      & ARC-Easy & 59.59 & 58.83 & \underline{61.29} & 58.48 & \textbf{62.81} \\
      & BoolQ & 58.92 & \textbf{60.71} & 57.96 & 58.37 & \underline{59.71} \\
      & CommonsenseQA & 37.37 & 37.37 & 35.54 & \underline{38.36} & \textbf{39.53} \\
      & COPA & 73.33 & \textbf{73.00} & 72.00 & 72.00 & \underline{72.67} \\
      & HellaSwag & 47.48 & 49.59 & \textbf{50.39} & 48.00 & \underline{50.13} \\
      & OpenBookQA & 32.20 & \textbf{35.47} & \underline{33.20} & 31.73 & 33.07 \\
      & PIQA & 70.44 & \underline{70.86} & 70.78 & 69.33 & \textbf{71.13} \\
      & SciQ & 85.37 & 84.67 & \textbf{87.40} & 85.10 & \underline{85.53} \\
      & SocialIQA & 43.36 & 43.36 & \textbf{44.23} & 43.88 & \underline{44.17} \\
      & WinoGrande & 54.91 & \textbf{56.14} & 55.25 & 53.33 & \underline{55.75} \\
      \cmidrule(lr){2-7}
      & Avg.\ Acc. & 51.90 & \underline{52.49} & 52.30 & 51.48 & \textbf{52.76} \\
    \bottomrule
  \end{tabular}
\end{table}

\FloatBarrier

Across three backbone scales, HiRes improves downstream accuracy over the vanilla
baseline by $0.24$, $0.39$, and $0.86$ percentage points from the smallest to
the largest model, achieving the highest aggregate accuracy among the compared
methods at the largest scale. Individual tasks favor different methods, but
HiRes delivers consistent aggregate gains and is competitive with other residual expansion methods.

\FloatBarrier
\Needspace{310pt}
\subsection{Long-Horizon MoE Pretraining}
\label{app:long-horizon-moe}

For the long-horizon experiment in
Section~\ref{sec:long-horizon-moe-pretraining},
Table~\ref{tab:moe-model-card} summarizes the MoE configuration and
500B-token results.

\begin{table}[!htbp]
  \caption{\textbf{On OLMoE-1B-7B, HiRes improves losses and most
  downstream tasks after 500B tokens.}
  \textbf{(a)} Model and training settings inherited from
  OLMoE~\citep{muennighoff2024olmoe}, plus the HiRes configuration.
  \textbf{(b)} Losses and 12-task downstream accuracy (\%).
  \textbf{Bold}: the better of the two.}
  \label{tab:moe-model-card}
  \centering
  \footnotesize
  \begin{minipage}[t]{0.445\textwidth}
    \vspace{0pt}
    \centering
    \textbf{(a) Model and training configuration}\\[3pt]
    \setlength{\tabcolsep}{5pt}
    \begin{tabular}{lr}
      \toprule
      Setting & Value \\
      \midrule
      Total parameters & 6.9B \\
      Active parameters & 1.3B \\
      Transformer layers & 16 \\
      Model width $d$ & 2,048 \\
      Attention heads & 16 \\
      Context length & 4,096 \\
      MoE experts & 64 \\
      MoE top-$k$ & 8 \\
      Expert hidden size & 2,048 \\
      Global batch size & 768 \\
      Peak learning rate & $4.0{\times}10^{-4}$ \\
      Warmup ratio & 2\% \\
      Training tokens & 500B \\
      \midrule
      HiRes outer width $D$ & 4,096 \\
      HiRes blocks $B$ & 4 \\
      HiRes added parameters & 83.9M \\
      Interface initialization & $\sigma(\alpha)=0.1$ \\
      \bottomrule
    \end{tabular}
  \end{minipage}\hfill
  \begin{minipage}[t]{0.535\textwidth}
    \vspace{0pt}
    \centering
    \textbf{(b) Training and downstream results}\\[3pt]
    \setlength{\tabcolsep}{5pt}
    \begin{tabular}{lrr}
      \toprule
      Metric & Baseline & HiRes ($n=2,B=4$) \\
      \midrule
      Train loss & 2.3663 & \textbf{2.3455} \\
      Val.\ loss & 2.3491 & \textbf{2.3311} \\
      \midrule
      MMLU & 41.58 & \textbf{42.19} \\
      ARC-Challenge & 44.70 & \textbf{49.61} \\
      ARC-Easy & 76.43 & \textbf{78.54} \\
      BoolQ & 66.39 & \textbf{70.09} \\
      CommonsenseQA & \textbf{49.88} & 49.66 \\
      COPA & \textbf{87.67} & 86.67 \\
      HellaSwag & 73.19 & \textbf{74.29} \\
      OpenBookQA & 44.33 & \textbf{44.80} \\
      PIQA & 78.65 & \textbf{79.67} \\
      SciQ & 93.13 & \textbf{93.60} \\
      SocialIQA & 47.78 & \textbf{48.43} \\
      WinoGrande & \textbf{67.17} & 67.01 \\
      \midrule
      Avg.\ Acc. & 64.24 & \textbf{65.38} \\
      \bottomrule
    \end{tabular}
  \end{minipage}
\end{table}

\FloatBarrier
\section{Detailed Ablation Results}

\subsection{Ablation on Block Count and Expansion Ratio}
\label{app:ablation-details}

Complementing Section~\ref{sec:ablation}, Table~\ref{tab:nbablation} reports the expansion-ratio and block-count ablations.

\begin{table}[!htbp]
  \caption{\textbf{Ablation results on expansion ratio $n$ and block count $B$.}
  Loss improves quickly up to $n=2$ and $B=4$ and slowly
  beyond. All runs use $(d,L)=(768,12)$ and 50B tokens.}
  \label{tab:nbablation}
  \centering
  \small
  \setlength{\tabcolsep}{7pt}
  \begin{tabular}{lrrr}
    \toprule
    Configuration & Params & Train loss & Val.\ loss \\
    \midrule
    \multicolumn{4}{l}{\emph{Expansion ratio at $B=4$}} \\
    $n=0.75$ & 142.9M & 3.2034 & 3.1735 \\
    $n=1$    & 144.3M & 3.1869 & 3.1573 \\
    $n=1.5$  & 147.3M & 3.1738 & 3.1450 \\
    $n=2$    & 150.2M & 3.1645 & 3.1300 \\
    $n=2.5$  & 153.2M & 3.1631 & 3.1306 \\
    $n=3$    & 156.1M & 3.1586 & 3.1242 \\
    $n=4$    & 162.0M & 3.1591 & 3.1251 \\
    \midrule
    \multicolumn{4}{l}{\emph{Block count at $n=2$}} \\
    $B=1$  & 143.2M & 3.2170 & 3.1877 \\
    $B=2$  & 145.5M & 3.1950 & 3.1658 \\
    $B=3$  & 147.9M & 3.1747 & 3.1426 \\
    $B=4$  & 150.2M & 3.1645 & 3.1300 \\
    $B=6$  & 155.0M & 3.1557 & 3.1249 \\
    $B=12$ & 169.1M & 3.1540 & 3.1172 \\
    \bottomrule
  \end{tabular}
\end{table}

\FloatBarrier
\Needspace{440pt}
\subsection{Architecture Ablations}
\label{app:supplementary-ablations}

Table~\ref{tab:supplementary-architecture-ablation} extends the architecture ablations in Section~\ref{sec:architecture-ablation} with interface, stabilization, write-rule, and transport variants.

\begin{table}[!htbp]
  \caption{\textbf{Supplementary interface ablations at $(d,L)=(768,12)$.}
  All models train for 50B tokens.}
  \label{tab:supplementary-architecture-ablation}
  \centering
  \small
  \setlength{\tabcolsep}{7pt}
  \renewcommand{\arraystretch}{0.96}
  \begin{tabular}{lcc}
    \toprule
    Variant & Train loss & Val.\ loss \\
    \midrule
    \multicolumn{3}{l}{\emph{Boundary normalization $\times$ outer residual}} \\
    Baseline & 3.2228 & 3.1941 \\
    Boundary RMSNorm only (HiNorm) & 3.1956 & 3.1714 \\
    Identity $R/W$ + outer residual & 3.2118 & 3.1807 \\
    Identity $R/W$ + outer residual + RMSNorm & 3.1943 & 3.1686 \\
    \midrule
    \multicolumn{3}{l}{\emph{Learned interface $\times$ scalar stabilization}} \\
    Learned $R/W$, stabilized & 3.1645 & 3.1300 \\
    Learned $R/W$, direct & 3.1686 & 3.1375 \\
    Frozen random $R/W$, stabilized & 3.1905 & 3.1614 \\
    Frozen random $R/W$, direct & 3.2181 & 3.1927 \\
    \midrule
    \multicolumn{3}{l}{\emph{Boundary RMSNorm $\times$ scalar stabilization}} \\
    With RMSNorm, stabilized & 3.1645 & 3.1300 \\
    Without RMSNorm, stabilized & 3.1874 & 3.1552 \\
    With RMSNorm, direct & 3.1686 & 3.1375 \\
    Without RMSNorm, direct & 3.1938 & 3.1591 \\
    \midrule
    \multicolumn{3}{l}{\emph{Write rule at $n=2,B=4$}} \\
    Full-output write & 3.1645 & 3.1300 \\
    Increment-only write & 3.1653 & 3.1341 \\
    \midrule
    \multicolumn{3}{l}{\emph{Read/write direction at $n=1,B=4$}} \\
    Complete read/write interface & 3.1869 & 3.1573 \\
    Read-only interface & 3.1902 & 3.1630 \\
    Write-only interface & 3.1897 & 3.1614 \\
    \bottomrule
  \end{tabular}
\end{table}

Boundary RMSNorm and the outer residual each improve over the baseline when
introduced separately, and their combination performs best in the
identity-interface factorial. Stabilization improves both learned and frozen interfaces, with the larger
gain for frozen random maps. Boundary RMSNorm yields a similar improvement
with and without stabilization, so its contribution is not contingent on the
scalar gates. The increment-only variant replaces the default write
$W_b\Phi_b(x_{b,0})$ with $W_b[\Phi_b(x_{b,0})-x_{b,0}]$, committing only the
update accumulated inside the block, so the outer skip is exactly the identity
rather than $I+W_bR_b$; its result is close to full-output writing. At equal residual and compute width, both one-sided variants
underperform the complete interface, indicating that learned transport in both
directions is preferable.

\Needspace{310pt}
\subsubsection{Boundary Normalization}
\label{app:boundary-normalization}

\begin{figure}[!htbp]
  \centering
  \begin{subfigure}[t]{0.49\textwidth}
    \centering
    \includegraphics[width=\linewidth]{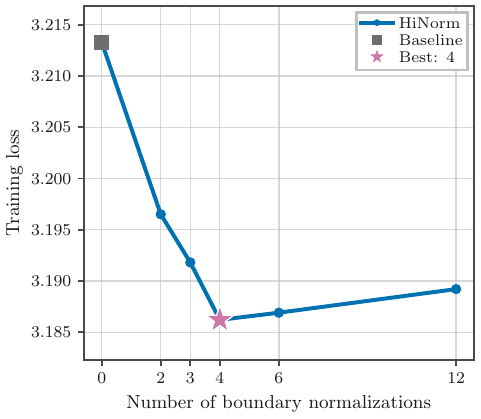}
    \caption{Boundary-frequency ablation.}
    \label{fig:hinorm-frequency}
  \end{subfigure}
  \hfill
  \begin{subfigure}[t]{0.49\textwidth}
    \centering
    \includegraphics[width=\linewidth]{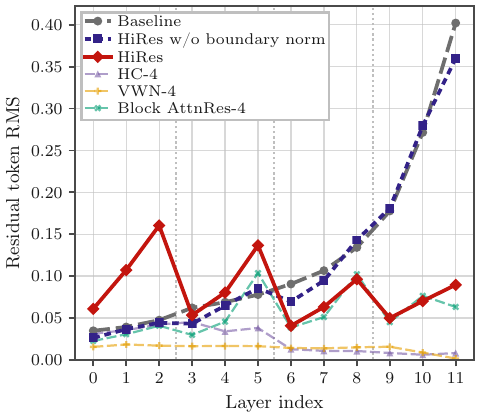}
    \caption{Per-layer residual activation scale.}
    \label{fig:activation-norm}
  \end{subfigure}
  \caption{\textbf{Boundary normalization lowers loss and regulates
  residual scale at $(d,L)=(768,12)$.}
  \textbf{(a)} HiNorm training loss under different boundary normalization frequencies.
  \textbf{(b)} Boundary normalization regulates residual accumulation as in other residual-expansion methods;
  dotted lines mark the HiRes block boundaries ($B=4$).}
  \label{fig:hinorm-analysis}
\end{figure}

Figure~\ref{fig:hinorm-analysis} shows the boundary-frequency ablation and the
per-layer residual scale discussed in Section~\ref{sec:design-choices}.

\subsubsection{Interface Stabilization Dynamics}
\label{sec:alpha}

\begin{figure}[!htbp]
  \centering
  \begin{subfigure}[t]{0.49\textwidth}
    \centering
    \includegraphics[width=\linewidth]{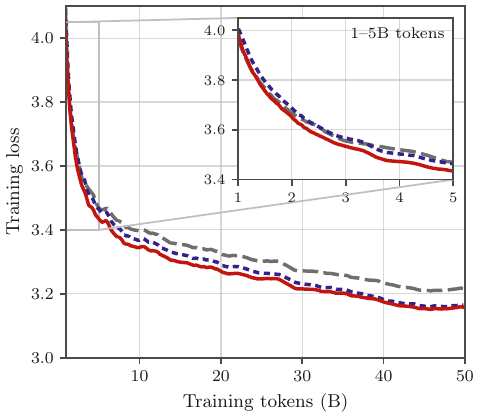}
    \caption{Training trajectories and early-token inset.}
    \label{fig:alpha-curves}
  \end{subfigure}
  \hfill
  \begin{subfigure}[t]{0.49\textwidth}
    \centering
    \includegraphics[width=\linewidth]{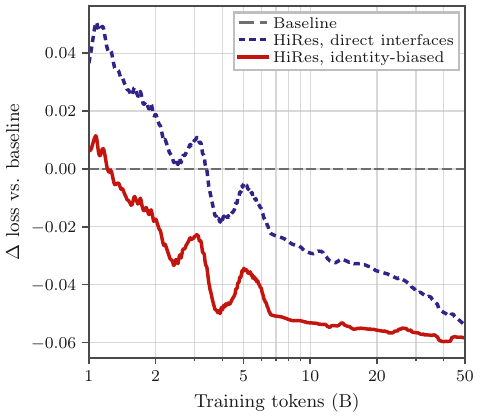}
    \caption{Loss difference from baseline.}
    \label{fig:alpha-difference}
  \end{subfigure}
  \caption{\textbf{Identity-biased initialization stabilizes early interface
  training.} Training loss of the baseline and of HiRes with direct or
  identity-biased interfaces under $(d,L)=(768,12)$.
  \textbf{(a)} Full training loss trajectories.
  \textbf{(b)} Loss difference from the baseline.
  Identity-biased interfaces start near baseline and maintain lower training loss than direct ones.}
  \label{fig:alpha-training}
\end{figure}

The scalar gates in Equation~\eqref{eq:interface-stabilization} bias each
read/write map toward the coordinate-aligned identity at initialization. We
compare the default initialization, $\sigma(\alpha)=0.1$, with direct dense
interfaces. Figure~\ref{fig:alpha-training} shows that direct interfaces
initially train worse than the baseline and only later cross below it. The
identity-biased parameterization starts closer to the baseline and remains
better throughout the early optimization phase. Stabilization therefore
prevents newly initialized interfaces from immediately disrupting residual
transport. The final results in
Table~\ref{tab:supplementary-architecture-ablation} further show that
stabilization is an optimization aid rather than a requirement for the
architecture.

\FloatBarrier
\section{Parameters, Compute, Memory, and I/O Costs}
\label{app:resource-accounting}

\paragraph{Parameters, compute, and memory.}
HiRes leaves all attention and FFN modules unchanged. Its additional
parameters come from $2B$ block interfaces, two endpoint projections, $B$ outer
RMSNorm, and $2B+2$ stabilization scalars:
\begin{equation}
  \Delta P
  =
  \underbrace{2BdD}_{R_b,W_b}
  +\underbrace{2dD}_{E,H}
  +\underbrace{BD}_{\text{Norm}}
  +\underbrace{(2B+2)}_{\text{scalar gates}} .
  \label{eq:parameter-overhead}
\end{equation}
For the OLMoE-1B-7B model in
Section~\ref{sec:long-horizon-moe-pretraining}, where
$(d,L,D,B)=(2048,16,4096,4)$, this amounts to
83.9M additional parameters, approximately
$6.5\%$ of the model's 1.3B active parameters and $1.2\%$ of its 6.9B total
parameters.
The dominant additional training compute comes from the forward and backward passes through the
$2(B+1)$ dense interface projections. Over $N_{\mathrm{train}}$ training tokens, this amounts to
\begin{equation}
  \Delta C_{\mathrm{train}}
  \approx 12 N_{\mathrm{train}}(B+1)dD
  \quad\text{FLOPs},
  \label{eq:training-compute}
\end{equation}
excluding the lower-order normalization and scalar-gate operations.
Relative to the vanilla residual baseline, HiRes additionally stores only the outer states at the
$(B+1)$ block boundaries. For a micro-batch of $N_{\mathrm{mb}}$ tokens, this
activation-memory overhead is
\begin{equation}
  \Delta A
  =N_{\mathrm{mb}}(B+1)D
  \quad\text{activation elements}.
  \label{eq:activation-state}
\end{equation}

\paragraph{Activation I/O.}
Let $f_l:\mathbb{R}^d\!\rightarrow\!\mathbb{R}^d$ denote one sublayer; an
$L$-layer Transformer has $2L$ sublayers. We count forward activation
elements per token, excluding parameter reads and $f_l$'s internal traffic.
For HiRes, the count includes its two endpoint maps.
Following AttnRes~\citep{kimiteam2026attentionresiduals}, an input constructed
only for $f_l$ is passed to it directly rather than written, whereas residual
states reused by later merges are written.
Table~\ref{tab:residual_io} summarizes the result for each method, and the
paragraphs below derive each row.

\begin{table}[!htbp]
  \centering
  \small
  \setlength{\tabcolsep}{4pt}
  \renewcommand{\arraystretch}{1.12}
  \caption{\textbf{HiRes reaches the lowest activation I/O among residual-expansion
  methods.} We compare optimized implementations for all methods and count residual-related activation reads and writes. $m$ is the HC/mHC/VWN expansion rate, $Q$ the number of
  AttnRes blocks, and $n=D/d$ and $B$ the HiRes
  expansion ratio and block count. The fourth column divides total I/O by the
  $2L$ sublayers; typical values use $L=16$, $Q=B=4$, $m=4$, and $n=2$.}
  \label{tab:residual_io}
  \resizebox{\textwidth}{!}{\begin{tabular}{@{}lcccc@{}}
    \toprule
    Method & Total reads & Total writes &
    \twolinehead{Average I/O}{per sublayer} & \twolinehead{Typical}{values} \\
    \midrule
    Baseline
      & $4Ld$ & $2Ld$ & $3d$ & $3d$ \\
    HC (fused)
      & $2L(3m{+}1)d$ & $2Lmd$ & $(4m{+}1)d$ & $17d$ \\
    mHC (fused)
      & $2L(3m{+}1)d$ & $2Lmd$ & $(4m{+}1)d$ & $17d$ \\
    VWN (fused)
      & $2L(3m{+}1)d$ & $2Lmd$ & $(4m{+}1)d$ & $17d$ \\
    Full AttnRes (two-phase)
      & $L\left(Q+\frac{2L}{Q}\right)d$ & $2Ld$
      & $\frac{1}{2}\left(Q+\frac{2L}{Q}+2\right)d$ & $7d$ \\
    Block AttnRes (two-phase)
      & $\left[6L+\frac{Q(Q-1)}{2}\right]d$ & $4Ld$
      & $\left[5+\frac{Q(Q-1)}{4L}\right]d$ & $5.2d$ \\
    \textbf{HiRes (fused)}
      & $\boldsymbol{[4L+B(2n{+}1)+n+1]d}$
      & $\boldsymbol{[2L+B(n{+}1)+n+1]d}$
      & $\boldsymbol{\left[3+\frac{B(3n+2)+2(n+1)}{2L}\right]d}$ & $\boldsymbol{4.2d}$ \\
    \bottomrule
  \end{tabular}}
\end{table}

\paragraph{Residual and HiRes.}
A vanilla merge moves $3d$ elements per sublayer, or $6Ld$ in total. HiRes retains
this cost for inner merges. At each block boundary, the read projection reads
$D$ elements and writes $d$ elements; fusing the write projection, outer
addition, and RMSNorm then reads $D+d$ elements and writes $D$ elements. Each
block therefore incurs $3D+2d$ elements of traffic, and the two endpoint maps
move another $2(D+d)$, giving the overhead
\begin{equation}
  \Delta\mathrm{IO}
  =
  B(3D+2d)+2(D+d)
  \quad\text{elements per token}
  \label{eq:io-overhead}
\end{equation}
and the total
\begin{equation}
  \mathrm{IO}_{\mathrm{HiRes}}
  =\left[6L+B(3n+2)+2(n+1)\right]d,\qquad n=D/d.
\end{equation}

\paragraph{HC, mHC and VWN.}
HC~\citep{zhu2025hyperconnections} maintains $m$ width-$d$ residual streams,
$X_l\in\mathbb{R}^{m\times d}$. Its three connections are generated from the
current state,
\begin{equation}
  \left(H_l^{\mathrm{pre}},H_l^{\mathrm{post}},
  H_l^{\mathrm{res}}\right)=G_l(X_l),
  \label{eq:hc-connections}
\end{equation}
with pre-connection $H_l^{\mathrm{pre}}\in\mathbb{R}^{1\times m}$,
post-connection $H_l^{\mathrm{post}}\in\mathbb{R}^{m\times 1}$, and a residual
connection $H_l^{\mathrm{res}}\in\mathbb{R}^{m\times m}$:
\begin{equation}
  x_l=H_l^{\mathrm{pre}}X_l,\qquad
  y_l=f_l(x_l),\qquad
  X_{l+1}=H_l^{\mathrm{res}}X_l+
  H_l^{\mathrm{post}}y_l .
  \label{eq:hc-recurrence}
\end{equation}

Following the dynamic HC parameterization, HC constructs three connections by
self-gated projections. Taking $H_l^{\mathrm{pre}}$ as an example:
\begin{equation}
  H_l^{\mathrm{pre}} = a_l^{\mathrm{pre}}\cdot \operatorname{tanh}(\operatorname{Norm}_l^{\mathrm{pre}}(X_l)W_l^{\mathrm{pre}})^\top + b_l^{\mathrm{pre}},
  \label{eq:hc-pre-connection}
\end{equation}
where $a_l^{\mathrm{pre}} \in \mathbb{R}$ is a scaling factor, $b_l^{\mathrm{pre}} \in \mathbb{R}^{1\times m}$ is a bias, and $W_l^{\mathrm{pre}} \in \mathbb{R}^{d\times 1}$ is the projection weight.

In mHC~\citep{xie2025mhc}, $G_l$ normalizes $\operatorname{vec}(X_l)$, jointly projects it to
$m^2+2m$ logits, applies sigmoid transforms to the pre/post logits, and applies
Sinkhorn--Knopp iterations to $H_l^{\mathrm{res}}$.

In VWN~\citep{seed2025vwn}, each stream is split into $r$ parts of width $d/r$, forming $X_l\in\mathbb{R}^{mr\times d/r}$.
The remaining operations are similar to HC.

The fused schedules below are our analytic constructions from these published
recurrences rather than traffic reported by the source papers. For each
method, we can pack the three projections in~\eqref{eq:hc-connections} and
fold~\eqref{eq:hc-pre-connection} into one kernel, reading $md$ elements once without
materializing the normalized and intermediate states. Because $H_l^{\mathrm{pre}}$ is available only after the projection
has reduced over all of $X_l$, the subsequent pre-connection in~\eqref{eq:hc-recurrence} makes a second
$md$ read and writes $d$. After $f_l$, the residual
connection and post-connection are fused into one kernel that reads $md+d$ and
writes $md$. Thus the activation traffic for each sublayer is
\begin{equation}
  R=(3m+1)d,\qquad W=md,\qquad \mathrm{IO}=(4m+1)d,
\end{equation}
and over the network,
\begin{equation}
  \mathrm{IO}_{\mathrm{HC}}=\mathrm{IO}_{\mathrm{mHC}}=\mathrm{IO}_{\mathrm{VWN}}
  =2L(4m+1)d .
\end{equation}
This omits the lower-order $m^2+2m$ coefficient traffic.

\paragraph{AttnRes.}
Full AttnRes~\citep{kimiteam2026attentionresiduals} stores every preceding layer output $v_i\in\mathbb{R}^{d}$ and
uses a learned query $q_l\in\mathbb{R}^{d}$ to retrieve across depth:
\begin{equation}
  \alpha_{i\rightarrow l}
  =\frac{\exp\!\left(q_l^\top\operatorname{RMSNorm}(v_i)\right)}
  {\sum_{j<l}\exp\!\left(q_l^\top\operatorname{RMSNorm}(v_j)\right)},
  \qquad
  h_l=\sum_{i<l}\alpha_{i\rightarrow l}v_i .
\end{equation}
Block AttnRes instead stores one completed representation per architecture
block and includes the current block's partial sum as an additional source.

For the two-phase schedule, partition the $2L$ sublayers into $Q$ execution
blocks of $S=2L/Q$ sublayers. Phase~1 batches the $S$ fixed queries in a block and
attends to completed history once, returning one $d$-wide output and
softmax statistics per sublayer. Phase~2 attends to the sequentially available
intra-block history and merges both phases using online
softmax~\citep{milakov2018online}. For Full
AttnRes, execution block $q\in\{0,\ldots,Q-1\}$ has $qS$ completed outputs.
Phase~1 reads these once for all $S$ queries, contributing
$qSd$ elements per block, and writes one $d$-wide result per sublayer.  Within
a block, sublayer $j$ reads $j$ earlier intra-block outputs in Phase~2, which
also reads back its Phase~1 result for the merge and passes $h_l$ directly to
$f_l$.  Summing both triangular histories over depth gives
\begin{equation}
  \begin{aligned}
  R_{\mathrm{FullAttnRes}}
    &=\left[S\sum_{q=0}^{Q-1}q
      +Q\sum_{j=0}^{S-1}j+2L\right]d
      =L(Q+S)d,\\
  W_{\mathrm{FullAttnRes}}&=2Ld,\\
  \overline{\mathrm{IO}}_{\mathrm{FullAttnRes}}
    &=\frac{Q+S+2}{2}d .
  \end{aligned}
\end{equation}
Here the two phases each write one $d$-wide result per sublayer.  Block AttnRes
instead reads only the $q$ completed block summaries in Phase~1.  Summing this
history gives $Q(Q-1)d/2$ reads; its fused partial-sum attention and
online-softmax merge additionally read $3d$ and write the $d$-wide partial sum
per sublayer.  Thus
\begin{equation}
  \begin{aligned}
  R_{\mathrm{BlockAttnRes}}
    &=\left[6L+\frac{Q(Q-1)}{2}\right]d,\\
  W_{\mathrm{BlockAttnRes}}&=4Ld,\\
  \overline{\mathrm{IO}}_{\mathrm{BlockAttnRes}}
    &=\left[5+\frac{Q-1}{2S}\right]d .
  \end{aligned}
\end{equation}

\section{Agent-Driven Optimization of Residual Mechanisms}
\label{app:systems}

To assess the practical efficiency and ease of adoption of each residual expansion mechanism, we measure both its final training
efficiency and the agent-driven optimization required to reach
it, comparing framework-level PyTorch and mechanism-specific Triton
implementations.

\subsection{Experimental Protocol}
We evaluate $(d_{\mathrm{model}},L)\in
\{(768,12),(1024,16),(1280,20)\}$ on eight GPUs using FSDP
\texttt{FULL\_SHARD}, bfloat16, sequence length $4096$, and per-GPU
micro-batch $4$. Every measurement includes forward, backward, and the
optimizer update; peak memory is reported per GPU. The attention, MLP,
normalization, optimizer, precision, and measurement harness are shared, and
only mechanism-specific code may change.

Starting from functionally equivalent naive implementations, independent
Claude Opus 5 agents receive the same prompt and optimize HiRes, VWN-4,
mHC-4, and Block AttnRes-4 in two stages: the PyTorch stage that may leverage graph
transformations, \texttt{torch.compile}, batching, and layout changes; and the
Triton stage that adds mechanism-specific fused kernels. Every candidate must match
the reference outputs and gradients at all three scales. Candidates are
measured against the current best and retained only if throughput
improves by at least $0.2\%$ at every scale. We report the best implementation from each stage.
The final optimized Triton implementation is used to measure the efficiency of each residual expansion mechanism in Table~\ref{tab:priorarttransfer}.

\subsection{Optimization Trajectory}

\begin{figure}[!htb]
  \centering
  \includegraphics[width=\linewidth]{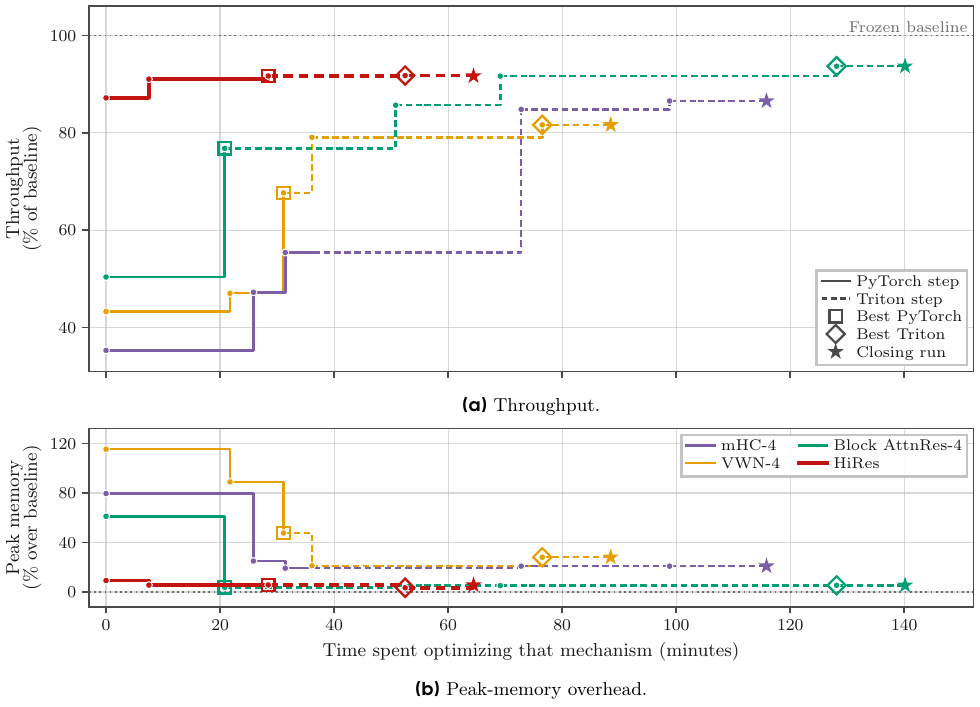}
  \caption{\textbf{Agent-driven optimization campaign.} \textbf{(a)} Throughput (\% of the frozen baseline). \textbf{(b)}
  Peak-memory overhead at $(d,L)=(768,12)$ versus optimization time. Dots are
  accepted improvements; solid and dashed segments are the PyTorch and Triton
  stages; square, diamond, and star mark the best PyTorch, best Triton, and
  closing implementations. HiRes reaches its best throughput early and without custom
  kernels.}
  \label{fig:sys-campaign}
\end{figure}

We visualize the optimization campaign in Figure~\ref{fig:sys-campaign}.
Three patterns emerge. First, HiRes maintains the highest throughput among the
residual-expansion methods throughout most of the campaign, until the fully
optimized Block AttnRes overtakes it near the end. Second, HiRes reaches its
best implementation in less optimization time than the other
methods in our campaign. Third, HiRes' PyTorch implementation already matches its final Triton
throughput, with custom kernels primarily reducing memory. Other methods instead rely on Triton for substantial throughput gains.

\begin{table}[!htbp]
  \caption{\textbf{Best PyTorch- and Triton-stage performance of each
  method.} Throughput and peak-memory overhead are relative to the frozen
  baseline. Within each stage, best/second-best values are
  \textbf{bold}/\underline{underlined}. HiRes leads in throughput and memory without custom Triton kernels.}
  \label{tab:sys-stage}
  \centering
  \small
  \setlength{\tabcolsep}{5pt}
  \begin{tabular}{llrrrrrr}
    \toprule
    & & \multicolumn{3}{c}{Throughput (\% of baseline)}
      & \multicolumn{3}{c}{Peak memory (\% over baseline)} \\
    \cmidrule(lr){3-5}\cmidrule(lr){6-8}
    Method & Stage & $(768,12)$ & $(1024,16)$ & $(1280,20)$
      & $(768,12)$ & $(1024,16)$ & $(1280,20)$ \\
    \midrule
    \multirow{2}{*}{mHC-4}
      & PyTorch & 58.1 & 56.4 & 55.2 & +19.1 & +25.2 & +29.7 \\
      & Triton  & 86.5 & 88.2 & 86.1 & +20.8 & +27.0 & +31.4 \\
    \midrule
    \multirow{2}{*}{VWN-4}
      & PyTorch & 68.3 & 69.3 & 69.2 & +47.6 & +58.7 & +66.2 \\
      & Triton  & 81.6 & 84.4 & 80.5 & +28.1 & +34.2 & +38.4 \\
    \midrule
    \multirow{2}{*}{Block AttnRes-4}
      & PyTorch & \underline{80.3} & \underline{77.4} & \underline{75.1}
        & \textbf{+3.7} & \textbf{+4.9} & \underline{+5.8} \\
      & Triton  & \textbf{93.7} & \textbf{94.9} & \textbf{94.5}
        & \underline{+5.3} & \underline{+6.9} & \underline{+8.0} \\
    \midrule
    \multirow{2}{*}{HiRes}
      & PyTorch & \textbf{91.7} & \textbf{92.7} & \textbf{93.0}
        & \underline{+5.5} & \underline{+5.7} & \textbf{+5.5} \\
      & Triton  & \underline{91.7} & \underline{92.8} & \underline{93.1}
        & \textbf{+3.4} & \textbf{+3.4} & \textbf{+3.3} \\
    \bottomrule
  \end{tabular}
\end{table}

\subsection{Practical Efficiency}
Table~\ref{tab:sys-stage} compares the best PyTorch-stage and fused Triton
implementations across scales. Without custom kernels, HiRes already achieves
$91.7$--$93.0\%$ of baseline throughput---the highest among the
residual-expansion methods---with only $5.5$--$5.7\%$ peak-memory overhead.
Triton fusion preserves this throughput while reducing the memory overhead to
$3.3$--$3.4\%$. The resulting implementation approaches the throughput of
Block AttnRes-4, exceeds that of mHC-4 and VWN-4, and has the lowest memory
overhead among the compared methods.

\FloatBarrier
\subsection{Latency Decomposition}
\begin{figure}[!htbp]
  \centering
  \includegraphics[width=\linewidth]{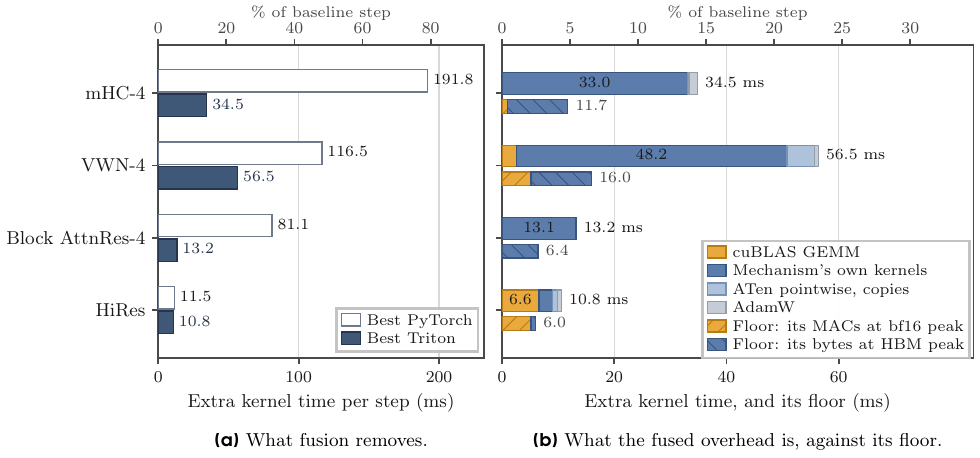}
  \caption{\textbf{Extra kernel time per step at $(1280,20)$ over the shared frozen
  backbone ($242.5$\,ms).} \textbf{(a)} Fusion (best Triton versus best
  PyTorch) cuts the overhead of the other methods by $2$--$6\times$ but leaves
  HiRes nearly unchanged. \textbf{(b)} Fused overhead by kernel type (upper
  bars) and its compute- and bandwidth-limited floors at the GPU's advertised
  bf16 and HBM peaks (lower bars): HiRes spends most of it in cuBLAS GEMMs, the
  others in their own mixing kernels.}
  \label{fig:sys-latency}
\end{figure}

Figure~\ref{fig:sys-latency} separates measured kernel time from analytic
compute- and bandwidth-limited floors.  HiRes spends most of its overhead in
the $2B+2$ dense interface GEMMs, whereas the comparison methods spend most of
theirs moving and mixing expanded states at every sublayer.  This explains why
fusion materially benefits the latter, while HiRes is already close to its
fused performance with framework-level optimizations.

\FloatBarrier
\Needspace{200pt}
\section{A Latent View}
\label{app:latent}

Besides expanding the residual state width, HiRes also admits a dual interpretation: relative to a $D$-dimensional
persistent hidden state, its attention and FFN computation occurs in a
$d$-dimensional latent space. Expanding residual width by $n=D/d$ is
therefore equivalent to compressing compute width by a factor of $n$; the wide state
preserves information across blocks, while the narrow state performs the
transformations within each block.

This view connects HiRes to module-specific latent designs. MLA compresses
the attention KV representation~\citep{deepseekai2024deepseekv2}, while
LatentMoE performs routed expert computation in a compressed
space~\citep{elango2026latentmoe}. HiRes extends the same separation between
state width and compute width to the residual hierarchy. Its key distinction
is interaction granularity: instead of exchanging information between hidden
and latent spaces at every sublayer, HiRes reads once at block entry and
writes once at block exit. Layer grouping therefore reduces hidden-latent
transport from per-sublayer to per-block while allowing computation to evolve
for multiple layers in the latent space.

\section{Limitations}
\label{app:limitations}

Our study has the following limitations.
First, HiRes decouples residual width from compute width, but we have not jointly
optimized these widths and model depth under a fixed compute budget.
The best allocation of capacity between carrying representations and
transforming them therefore remains unclear. Second, we use equal-sized
blocks, so exchanges between the wide state and the inner computation
occur at regular intervals. We have not examined whether different parts
of the network benefit from more frequent access to the wide state. Finally,
our results show the benefits of residual expansion, yet how the wider
residual state improves learning remains an open question. In particular,
what information is preserved across blocks and how individual sublayers
contribute to its transformation and reuse are not yet well understood.


\clearpage
\begin{thebibliography}{61}
\providecommand{\natexlab}[1]{#1}
\providecommand{\url}[1]{\texttt{#1}}
\expandafter\ifx\csname urlstyle\endcsname\relax
  \providecommand{\doi}[1]{doi: #1}\else
  \providecommand{\doi}{doi: \begingroup \urlstyle{rm}\Url}\fi

\bibitem[Ainslie et~al.(2023)Ainslie, Lee-Thorp, de~Jong, Zemlyanskiy, Lebron,
  and Sanghai]{ainslie2023gqa}
Joshua Ainslie, James Lee-Thorp, Michiel de~Jong, Yury Zemlyanskiy, Federico
  Lebron, and Sumit Sanghai.
\newblock {GQA}: Training generalized multi-query transformer models from
  multi-head checkpoints.
\newblock In Houda Bouamor, Juan Pino, and Kalika Bali, editors,
  \emph{Proceedings of the 2023 Conference on Empirical Methods in Natural
  Language Processing}, pages 4895--4901, Singapore, December 2023. Association
  for Computational Linguistics.
\newblock \doi{10.18653/v1/2023.emnlp-main.298}.

\bibitem[Ba et~al.(2016)Ba, Kiros, and Hinton]{ba2016layernorm}
Jimmy~Lei Ba, Jamie~Ryan Kiros, and Geoffrey~E. Hinton.
\newblock Layer normalization, 2016.

\bibitem[Bachlechner et~al.(2021)Bachlechner, Majumder, Mao, Cottrell, and
  McAuley]{bachlechner2020rezero}
Thomas Bachlechner, Bodhisattwa~Prasad Majumder, Henry Mao, Gary Cottrell, and
  Julian McAuley.
\newblock Rezero is all you need: fast convergence at large depth.
\newblock In Cassio de~Campos and Marloes~H. Maathuis, editors,
  \emph{Proceedings of the Thirty-Seventh Conference on Uncertainty in
  Artificial Intelligence}, volume 161 of \emph{Proceedings of Machine Learning
  Research}, pages 1352--1361. PMLR, 27--30 Jul 2021.

\bibitem[Baykal et~al.(2023)Baykal, Cutler, Dikkala, Ghosh, Panigrahy, and
  Wang]{baykal2023altup}
Cenk Baykal, Dylan Cutler, Nishanth Dikkala, Nikhil Ghosh, Rina Panigrahy, and
  Xin Wang.
\newblock Alternating updates for efficient transformers.
\newblock In A.~Oh, T.~Naumann, A.~Globerson, K.~Saenko, M.~Hardt, and
  S.~Levine, editors, \emph{Advances in Neural Information Processing Systems},
  volume~36, pages 76718--76736. Curran Associates, Inc., 2023.
\newblock \doi{10.52202/075280-3352}.

\bibitem[Bisk et~al.(2020)Bisk, Zellers, Le~bras, Gao, and Choi]{bisk2020piqa}
Yonatan Bisk, Rowan Zellers, Ronan Le~bras, Jianfeng Gao, and Yejin Choi.
\newblock Piqa: Reasoning about physical commonsense in natural language.
\newblock \emph{Proceedings of the AAAI Conference on Artificial Intelligence},
  34\penalty0 (05):\penalty0 7432--7439, April 2020.
\newblock ISSN 2159-5399.
\newblock \doi{10.1609/aaai.v34i05.6239}.

\bibitem[Clark et~al.(2019)Clark, Lee, Chang, Kwiatkowski, Collins, and
  Toutanova]{clark2019boolq}
Christopher Clark, Kenton Lee, Ming-Wei Chang, Tom Kwiatkowski, Michael
  Collins, and Kristina Toutanova.
\newblock {B}ool{Q}: Exploring the surprising difficulty of natural yes/no
  questions.
\newblock In Jill Burstein, Christy Doran, and Thamar Solorio, editors,
  \emph{Proceedings of the 2019 Conference of the North {A}merican Chapter of
  the Association for Computational Linguistics: Human Language Technologies,
  Volume 1 (Long and Short Papers)}, pages 2924--2936, Minneapolis, Minnesota,
  June 2019. Association for Computational Linguistics.
\newblock \doi{10.18653/v1/N19-1300}.

\bibitem[Clark et~al.(2018)Clark, Cowhey, Etzioni, Khot, Sabharwal, Schoenick,
  and Tafjord]{clark2018arc}
Peter Clark, Isaac Cowhey, Oren Etzioni, Tushar Khot, Ashish Sabharwal, Carissa
  Schoenick, and Oyvind Tafjord.
\newblock Think you have solved question answering? try arc, the ai2 reasoning
  challenge, 2018.

\bibitem[{DeepSeek-AI} et~al.(2024){DeepSeek-AI}, Liu, Feng, Wang, Wang, Liu,
  Zhao, Dengr, Ruan, Dai, Guo, Yang, Chen, Ji, Li, Lin, Luo, Hao, Chen, Li,
  Zhang, Xu, Yang, Zhang, Ding, Xin, Gao, Li, Qu, Cai, Liang, Guo, Ni, Li,
  Chen, Yuan, Qiu, Song, Dong, Gao, Guan, Wang, Zhang, Xu, Xia, Zhao, Zhang,
  Li, Wang, Zhang, Zhang, Tang, Li, Tian, Huang, Wang, Zhang, Zhu, Chen, Du,
  Chen, Jin, Ge, Pan, Xu, Chen, Li, Lu, Zhou, Chen, Wu, Ye, Ma, Wang, Zhou, Yu,
  Zhou, Zheng, Wang, Pei, Yuan, Sun, Xiao, Zeng, An, Liu, Liang, Gao, Zhang,
  Li, Jin, Wang, Bi, Liu, Wang, Shen, Chen, Chen, Nie, Sun, Wang, Liu, Xie, Yu,
  Song, Zhou, Yang, Lu, Su, Wu, Li, Wei, Zhu, Xu, Huang, Li, Zhao, Sun, Li,
  Wang, Zheng, Zhang, Xiong, Zhao, He, Tang, Piao, Dong, Tan, Liu, Wang, Guo,
  Zhu, Wang, Zou, Zha, Ma, Yan, You, Liu, Ren, Ren, Sha, Fu, Huang, Zhang, Xie,
  Hao, Shao, Wen, Xu, Zhang, Li, Wang, Gu, Li, and
  Xie]{deepseekai2024deepseekv2}
{DeepSeek-AI}, Aixin Liu, Bei Feng, Bin Wang, Bingxuan Wang, Bo~Liu, Chenggang
  Zhao, Chengqi Dengr, Chong Ruan, Damai Dai, Daya Guo, Dejian Yang, Deli Chen,
  Dongjie Ji, Erhang Li, Fangyun Lin, Fuli Luo, Guangbo Hao, Guanting Chen,
  Guowei Li, H.~Zhang, Hanwei Xu, Hao Yang, Haowei Zhang, Honghui Ding, Huajian
  Xin, Huazuo Gao, Hui Li, Hui Qu, J.~L. Cai, Jian Liang, Jianzhong Guo, Jiaqi
  Ni, Jiashi Li, Jin Chen, Jingyang Yuan, Junjie Qiu, Junxiao Song, Kai Dong,
  Kaige Gao, Kang Guan, Lean Wang, Lecong Zhang, Lei Xu, Leyi Xia, Liang Zhao,
  Liyue Zhang, Meng Li, Miaojun Wang, Mingchuan Zhang, Minghua Zhang, Minghui
  Tang, Mingming Li, Ning Tian, Panpan Huang, Peiyi Wang, Peng Zhang, Qihao
  Zhu, Qinyu Chen, Qiushi Du, R.~J. Chen, R.~L. Jin, Ruiqi Ge, Ruizhe Pan,
  Runxin Xu, Ruyi Chen, S.~S. Li, Shanghao Lu, Shangyan Zhou, Shanhuang Chen,
  Shaoqing Wu, Shengfeng Ye, Shirong Ma, Shiyu Wang, Shuang Zhou, Shuiping Yu,
  Shunfeng Zhou, Size Zheng, T.~Wang, Tian Pei, Tian Yuan, Tianyu Sun, W.~L.
  Xiao, Wangding Zeng, Wei An, Wen Liu, Wenfeng Liang, Wenjun Gao, Wentao
  Zhang, X.~Q. Li, Xiangyue Jin, Xianzu Wang, Xiao Bi, Xiaodong Liu, Xiaohan
  Wang, Xiaojin Shen, Xiaokang Chen, Xiaosha Chen, Xiaotao Nie, Xiaowen Sun,
  Xiaoxiang Wang, Xin Liu, Xin Xie, Xingkai Yu, Xinnan Song, Xinyi Zhou, Xinyu
  Yang, Xuan Lu, Xuecheng Su, Y.~Wu, Y.~K. Li, Y.~X. Wei, Y.~X. Zhu, Yanhong
  Xu, Yanping Huang, Yao Li, Yao Zhao, Yaofeng Sun, Yaohui Li, Yaohui Wang,
  Yi~Zheng, Yichao Zhang, Yiliang Xiong, Yilong Zhao, Ying He, Ying Tang, Yishi
  Piao, Yixin Dong, Yixuan Tan, Yiyuan Liu, Yongji Wang, Yongqiang Guo, Yuchen
  Zhu, Yuduan Wang, Yuheng Zou, Yukun Zha, Yunxian Ma, Yuting Yan, Yuxiang You,
  Yuxuan Liu, Z.~Z. Ren, Zehui Ren, Zhangli Sha, Zhe Fu, Zhen Huang, Zhen
  Zhang, Zhenda Xie, Zhewen Hao, Zhihong Shao, Zhiniu Wen, Zhipeng Xu, Zhongyu
  Zhang, Zhuoshu Li, Zihan Wang, Zihui Gu, Zilin Li, and Ziwei Xie.
\newblock Deepseek-v2: A strong, economical, and efficient mixture-of-experts
  language model, 2024.

\bibitem[Dehghani et~al.(2023)Dehghani, Djolonga, Mustafa, Padlewski, Heek,
  Gilmer, Steiner, Caron, Geirhos, Alabdulmohsin, Jenatton, Beyer, Tschannen,
  Arnab, Wang, Riquelme~Ruiz, Minderer, Puigcerver, Evci, Kumar, Steenkiste,
  Elsayed, Mahendran, Yu, Oliver, Huot, Bastings, Collier, Gritsenko, Birodkar,
  Vasconcelos, Tay, Mensink, Kolesnikov, Pavetic, Tran, Kipf, Lucic, Zhai,
  Keysers, Harmsen, and Houlsby]{dehghani2023vit22b}
Mostafa Dehghani, Josip Djolonga, Basil Mustafa, Piotr Padlewski, Jonathan
  Heek, Justin Gilmer, Andreas~Peter Steiner, Mathilde Caron, Robert Geirhos,
  Ibrahim Alabdulmohsin, Rodolphe Jenatton, Lucas Beyer, Michael Tschannen,
  Anurag Arnab, Xiao Wang, Carlos Riquelme~Ruiz, Matthias Minderer, Joan
  Puigcerver, Utku Evci, Manoj Kumar, Sjoerd~Van Steenkiste, Gamaleldin~Fathy
  Elsayed, Aravindh Mahendran, Fisher Yu, Avital Oliver, Fantine Huot, Jasmijn
  Bastings, Mark Collier, Alexey~A. Gritsenko, Vighnesh Birodkar,
  Cristina~Nader Vasconcelos, Yi~Tay, Thomas Mensink, Alexander Kolesnikov,
  Filip Pavetic, Dustin Tran, Thomas Kipf, Mario Lucic, Xiaohua Zhai, Daniel
  Keysers, Jeremiah~J. Harmsen, and Neil Houlsby.
\newblock Scaling vision transformers to 22 billion parameters.
\newblock In Andreas Krause, Emma Brunskill, Kyunghyun Cho, Barbara Engelhardt,
  Sivan Sabato, and Jonathan Scarlett, editors, \emph{Proceedings of the 40th
  International Conference on Machine Learning}, volume 202 of
  \emph{Proceedings of Machine Learning Research}, pages 7480--7512. PMLR,
  23--29 Jul 2023.

\bibitem[Elango et~al.(2026)Elango, Bhatia, Waleffe, Shafipour, Asida, Khattar,
  Assaf, Golub, Guman, Mitra, Zhao, Borkar, Zilberstein, Patwary, Shoeybi, and
  Rouhani]{elango2026latentmoe}
Venmugil Elango, Nidhi Bhatia, Roger Waleffe, Rasoul Shafipour, Tomer Asida,
  Abhinav Khattar, Nave Assaf, Maximilian Golub, Joey Guman, Tiyasa Mitra,
  Ritchie Zhao, Ritika Borkar, Ran Zilberstein, Mostofa Patwary, Mohammad
  Shoeybi, and Bita Rouhani.
\newblock Latentmoe: Toward optimal accuracy per flop and parameter in mixture
  of experts, 2026.

\bibitem[Fang et~al.(2023)Fang, CAI, Chen, Zhao, Tian, and Li]{fang2023mrla}
Yanwen Fang, Yuxi CAI, Jintai Chen, Jingyu Zhao, Guangjian Tian, and Guodong
  Li.
\newblock Cross-layer retrospective retrieving via layer attention.
\newblock In \emph{The Eleventh International Conference on Learning
  Representations}, 2023.

\bibitem[Fedus et~al.(2022)Fedus, Zoph, and Shazeer]{fedus2022switch}
William Fedus, Barret Zoph, and Noam Shazeer.
\newblock Switch transformers: Scaling to trillion parameter models with simple
  and efficient sparsity.
\newblock \emph{Journal of Machine Learning Research}, 23\penalty0
  (120):\penalty0 1--39, 2022.

\bibitem[Gao et~al.(2020)Gao, Biderman, Black, Golding, Hoppe, Foster, Phang,
  He, Thite, Nabeshima, Presser, and Leahy]{gao2020pile}
Leo Gao, Stella Biderman, Sid Black, Laurence Golding, Travis Hoppe, Charles
  Foster, Jason Phang, Horace He, Anish Thite, Noa Nabeshima, Shawn Presser,
  and Connor Leahy.
\newblock The pile: An 800gb dataset of diverse text for language modeling,
  2020.

\bibitem[He et~al.(2016)He, Zhang, Ren, and Sun]{he2016deep}
Kaiming He, Xiangyu Zhang, Shaoqing Ren, and Jian Sun.
\newblock Deep residual learning for image recognition.
\newblock In \emph{Proceedings of the IEEE Conference on Computer Vision and
  Pattern Recognition (CVPR)}, June 2016.

\bibitem[Hendrycks et~al.(2021)Hendrycks, Burns, Basart, Zou, Mazeika, Song,
  and Steinhardt]{hendrycks2021mmlu}
Dan Hendrycks, Collin Burns, Steven Basart, Andy Zou, Mantas Mazeika, Dawn
  Song, and Jacob Steinhardt.
\newblock Measuring massive multitask language understanding.
\newblock In \emph{International Conference on Learning Representations}, 2021.

\bibitem[Hoffmann et~al.(2022)Hoffmann, Borgeaud, Mensch, Buchatskaya, Cai,
  Rutherford, de~Las~Casas, Hendricks, Welbl, Clark, Hennigan, Noland,
  Millican, van~den Driessche, Damoc, Guy, Osindero, Simonyan, Elsen, Vinyals,
  Rae, and Sifre]{hoffmann2022chinchilla}
Jordan Hoffmann, Sebastian Borgeaud, Arthur Mensch, Elena Buchatskaya, Trevor
  Cai, Eliza Rutherford, Diego de~Las~Casas, Lisa~Anne Hendricks, Johannes
  Welbl, Aidan Clark, Thomas Hennigan, Eric Noland, Katherine Millican, George
  van~den Driessche, Bogdan Damoc, Aurelia Guy, Simon Osindero, Kar\'{e}n
  Simonyan, Erich Elsen, Oriol Vinyals, Jack Rae, and Laurent Sifre.
\newblock An empirical analysis of compute-optimal large language model
  training.
\newblock In S.~Koyejo, S.~Mohamed, A.~Agarwal, D.~Belgrave, K.~Cho, and A.~Oh,
  editors, \emph{Advances in Neural Information Processing Systems}, volume~35,
  pages 30016--30030. Curran Associates, Inc., 2022.
\newblock \doi{10.52202/068431-2176}.

\bibitem[Huang et~al.(2017)Huang, Liu, van~der Maaten, and
  Weinberger]{huang2017densely}
Gao Huang, Zhuang Liu, Laurens van~der Maaten, and Kilian~Q. Weinberger.
\newblock Densely connected convolutional networks.
\newblock In \emph{Proceedings of the IEEE Conference on Computer Vision and
  Pattern Recognition (CVPR)}, July 2017.

\bibitem[Kaplan et~al.(2020)Kaplan, McCandlish, Henighan, Brown, Chess, Child,
  Gray, Radford, Wu, and Amodei]{kaplan2020scaling}
Jared Kaplan, Sam McCandlish, Tom Henighan, Tom~B. Brown, Benjamin Chess, Rewon
  Child, Scott Gray, Alec Radford, Jeffrey Wu, and Dario Amodei.
\newblock Scaling laws for neural language models, 2020.

\bibitem[Kim et~al.(2025)Kim, Lee, Park, Oh, Kim, Yoo, Shin, Han, Shin, and
  Yoo]{kim2025periln}
Jeonghoon Kim, Byeongchan Lee, Cheonbok Park, Yeontaek Oh, Beomjun Kim, Taehwan
  Yoo, Seongjin Shin, Dongyoon Han, Jinwoo Shin, and Kang~Min Yoo.
\newblock Peri-{LN}: Revisiting normalization layer in the transformer
  architecture.
\newblock In Aarti Singh, Maryam Fazel, Daniel Hsu, Simon Lacoste-Julien, Felix
  Berkenkamp, Tegan Maharaj, Kiri Wagstaff, and Jerry Zhu, editors,
  \emph{Proceedings of the 42nd International Conference on Machine Learning},
  volume 267 of \emph{Proceedings of Machine Learning Research}, pages
  30400--30436. PMLR, 13--19 Jul 2025.

\bibitem[{Kimi Team} et~al.(2026){Kimi Team}, Chen, Zhang, Su, Xu, Pan, Wang,
  Wang, Chen, Yin, Chen, Yan, Wei, Zhang, Meng, Hong, Xie, Liu, Lu, Tai, Chen,
  Men, Guo, Charles, Lu, Sui, Zhu, Zhou, He, Huang, Xu, Wang, Lai, Du, Wu,
  Yang, and Zhou]{kimiteam2026attentionresiduals}
{Kimi Team}, Guangyu Chen, Yu~Zhang, Jianlin Su, Weixin Xu, Siyuan Pan, Yaoyu
  Wang, Yucheng Wang, Guanduo Chen, Bohong Yin, Yutian Chen, Junjie Yan, Ming
  Wei, Y.~Zhang, Fanqing Meng, Chao Hong, Xiaotong Xie, Shaowei Liu, Enzhe Lu,
  Yunpeng Tai, Yanru Chen, Xin Men, Haiqing Guo, Y.~Charles, Haoyu Lu, Lin Sui,
  Jinguo Zhu, Zaida Zhou, Weiran He, Weixiao Huang, Xinran Xu, Yuzhi Wang,
  Guokun Lai, Yulun Du, Yuxin Wu, Zhilin Yang, and Xinyu Zhou.
\newblock Attention residuals, 2026.

\bibitem[Li et~al.(2025)Li, Yin, and Liu]{li2025mixln}
Pengxiang Li, Lu~Yin, and Shiwei Liu.
\newblock Mix-ln: Unleashing the power of deeper layers by combining pre-ln and
  post-ln.
\newblock In Y.~Yue, A.~Garg, N.~Peng, F.~Sha, and R.~Yu, editors,
  \emph{International Conference on Learning Representations}, volume 2025,
  pages 98244--98258, 2025.

\bibitem[Li et~al.(2026)Li, Han, Cao, Huang, Zhou, Chen, Zhao, Jiang, Jiang,
  and Huang]{li2026siamesenorm}
Tianyu Li, Dongchen Han, Zixuan Cao, Haofeng Huang, Mengyu Zhou, Ming Chen,
  Erchao Zhao, Xiaoxi Jiang, Guanjun Jiang, and Gao Huang.
\newblock Siamesenorm: Breaking the barrier to reconciling pre/post-norm, 2026.

\bibitem[Lo et~al.(2020)Lo, Wang, Neumann, Kinney, and Weld]{lo2020s2orc}
Kyle Lo, Lucy~Lu Wang, Mark Neumann, Rodney Kinney, and Daniel Weld.
\newblock {S}2{ORC}: The semantic scholar open research corpus.
\newblock In Dan Jurafsky, Joyce Chai, Natalie Schluter, and Joel Tetreault,
  editors, \emph{Proceedings of the 58th Annual Meeting of the Association for
  Computational Linguistics}, pages 4969--4983, Online, July 2020. Association
  for Computational Linguistics.
\newblock \doi{10.18653/v1/2020.acl-main.447}.

\bibitem[Loshchilov and Hutter(2019)]{loshchilov2019adamw}
Ilya Loshchilov and Frank Hutter.
\newblock Decoupled weight decay regularization.
\newblock In \emph{International Conference on Learning Representations}, 2019.

\bibitem[Magnusson et~al.(2024)Magnusson, Bhagia, Hofmann, Soldaini, Jha,
  Tafjord, Schwenk, Walsh, Elazar, Lo, Groeneveld, Beltagy, Hajishirzi, Smith,
  Richardson, and Dodge]{magnusson2024paloma}
Ian Magnusson, Akshita Bhagia, Valentin Hofmann, Luca Soldaini, Ananya~Harsh
  Jha, Oyvind Tafjord, Dustin Schwenk, Evan Walsh, Yanai Elazar, Kyle Lo, Dirk
  Groeneveld, Iz~Beltagy, Hannaneh Hajishirzi, Noah~A. Smith, Kyle Richardson,
  and Jesse Dodge.
\newblock Paloma: A benchmark for evaluating language model fit.
\newblock In A.~Globerson, L.~Mackey, D.~Belgrave, A.~Fan, U.~Paquet,
  J.~Tomczak, and C.~Zhang, editors, \emph{Advances in Neural Information
  Processing Systems}, volume~37, pages 64338--64376. Curran Associates, Inc.,
  2024.
\newblock \doi{10.52202/079017-2052}.

\bibitem[Menghani et~al.(2025)Menghani, Kumar, and Kumar]{menghani2025laurel}
Gaurav Menghani, Ravi Kumar, and Sanjiv Kumar.
\newblock {LA}u{R}e{L}: Learned augmented residual layer.
\newblock In Aarti Singh, Maryam Fazel, Daniel Hsu, Simon Lacoste-Julien, Felix
  Berkenkamp, Tegan Maharaj, Kiri Wagstaff, and Jerry Zhu, editors,
  \emph{Proceedings of the 42nd International Conference on Machine Learning},
  volume 267 of \emph{Proceedings of Machine Learning Research}, pages
  43826--43836. PMLR, 13--19 Jul 2025.

\bibitem[Merity et~al.(2017)Merity, Xiong, Bradbury, and
  Socher]{merity2017pointer}
Stephen Merity, Caiming Xiong, James Bradbury, and Richard Socher.
\newblock Pointer sentinel mixture models.
\newblock In \emph{International Conference on Learning Representations}, 2017.

\bibitem[Mihaylov et~al.(2018)Mihaylov, Clark, Khot, and
  Sabharwal]{mihaylov2018openbookqa}
Todor Mihaylov, Peter Clark, Tushar Khot, and Ashish Sabharwal.
\newblock Can a suit of armor conduct electricity? a new dataset for open book
  question answering.
\newblock In Ellen Riloff, David Chiang, Julia Hockenmaier, and Jun{'}ichi
  Tsujii, editors, \emph{Proceedings of the 2018 Conference on Empirical
  Methods in Natural Language Processing}, pages 2381--2391, Brussels, Belgium,
  October-November 2018. Association for Computational Linguistics.
\newblock \doi{10.18653/v1/D18-1260}.

\bibitem[Milakov and Gimelshein(2018)]{milakov2018online}
Maxim Milakov and Natalia Gimelshein.
\newblock Online normalizer calculation for softmax, 2018.

\bibitem[Muennighoff et~al.(2025)Muennighoff, Soldaini, Groeneveld, Lo,
  Morrison, Min, Shi, Walsh, Tafjord, Lambert, Gu, Arora, Bhagia, Schwenk,
  Wadden, Wettig, Hui, Dettmers, Kiela, Farhadi, Smith, Koh, Singh, and
  Hajishirzi]{muennighoff2024olmoe}
Niklas Muennighoff, Luca Soldaini, Dirk Groeneveld, Kyle Lo, Jacob Morrison,
  Sewon Min, Weijia Shi, Pete Walsh, Oyvind Tafjord, Nathan Lambert, Yuling Gu,
  Shane Arora, Akshita Bhagia, Dustin Schwenk, David Wadden, Alexander Wettig,
  Binyuan Hui, Tim Dettmers, Douwe Kiela, Ali Farhadi, Noah Smith, Pang~Wei
  Koh, Amanpreet Singh, and Hanna Hajishirzi.
\newblock Olmoe: Open mixture-of-experts language models.
\newblock In Y.~Yue, A.~Garg, N.~Peng, F.~Sha, and R.~Yu, editors,
  \emph{International Conference on Learning Representations}, volume 2025,
  pages 62061--62121, 2025.

\bibitem[Nguyen and Salazar(2019)]{nguyen2019transformers}
Toan~Q. Nguyen and Julian Salazar.
\newblock Transformers without tears: Improving the normalization of
  self-attention.
\newblock In Jan Niehues, Rolando Cattoni, Sebastian St{\"u}ker, Matteo Negri,
  Marco Turchi, Thanh-Le Ha, Elizabeth Salesky, Ramon Sanabria, Loic Barrault,
  Lucia Specia, and Marcello Federico, editors, \emph{Proceedings of the 16th
  International Conference on Spoken Language Translation}, Hong Kong, November
  2-3 2019. Association for Computational Linguistics.

\bibitem[Pagliardini et~al.(2024)Pagliardini, Mohtashami, Fleuret, and
  Jaggi]{pagliardini2024denseformer}
Matteo Pagliardini, Amirkeivan Mohtashami, Francois Fleuret, and Martin Jaggi.
\newblock Denseformer: Enhancing information flow in transformers via depth
  weighted averaging.
\newblock In A.~Globerson, L.~Mackey, D.~Belgrave, A.~Fan, U.~Paquet,
  J.~Tomczak, and C.~Zhang, editors, \emph{Advances in Neural Information
  Processing Systems}, volume~37, pages 136479--136508. Curran Associates,
  Inc., 2024.
\newblock \doi{10.52202/079017-4336}.

\bibitem[Raffel et~al.(2020)Raffel, Shazeer, Roberts, Lee, Narang, Matena,
  Zhou, Li, and Liu]{raffel2020t5}
Colin Raffel, Noam Shazeer, Adam Roberts, Katherine Lee, Sharan Narang, Michael
  Matena, Yanqi Zhou, Wei Li, and Peter~J. Liu.
\newblock Exploring the limits of transfer learning with a unified text-to-text
  transformer.
\newblock \emph{Journal of Machine Learning Research}, 21\penalty0
  (140):\penalty0 1--67, 2020.

\bibitem[Reid et~al.(2022)Reid, Zhong, Gururangan, and
  Zettlemoyer]{reid2022m2d2}
Machel Reid, Victor Zhong, Suchin Gururangan, and Luke Zettlemoyer.
\newblock {M}2{D}2: A massively multi-domain language modeling dataset.
\newblock In Yoav Goldberg, Zornitsa Kozareva, and Yue Zhang, editors,
  \emph{Proceedings of the 2022 Conference on Empirical Methods in Natural
  Language Processing}, pages 964--975, Abu Dhabi, United Arab Emirates,
  December 2022. Association for Computational Linguistics.
\newblock \doi{10.18653/v1/2022.emnlp-main.63}.

\bibitem[Roemmele et~al.(2011)Roemmele, Bejan, and Gordon]{roemmele2011copa}
Melissa Roemmele, Cosmin~Adrian Bejan, and Andrew~S. Gordon.
\newblock Choice of plausible alternatives: An evaluation of commonsense causal
  reasoning.
\newblock In \emph{Logical Formalizations of Commonsense Reasoning: Papers from
  the AAAI Spring Symposium}, volume SS-11-06 of \emph{AAAI Technical Report},
  pages 90--95, 2011.

\bibitem[Sakaguchi et~al.(2021)Sakaguchi, Le~Bras, Bhagavatula, and
  Choi]{sakaguchi2021winogrande}
Keisuke Sakaguchi, Ronan Le~Bras, Chandra Bhagavatula, and Yejin Choi.
\newblock Winogrande: an adversarial winograd schema challenge at scale.
\newblock \emph{Communications of the ACM}, 64\penalty0 (9):\penalty0 99--106,
  August 2021.
\newblock ISSN 1557-7317.
\newblock \doi{10.1145/3474381}.

\bibitem[Sap et~al.(2019)Sap, Rashkin, Chen, Le~Bras, and
  Choi]{sap2019socialiqa}
Maarten Sap, Hannah Rashkin, Derek Chen, Ronan Le~Bras, and Yejin Choi.
\newblock Social {IQ}a: Commonsense reasoning about social interactions.
\newblock In Kentaro Inui, Jing Jiang, Vincent Ng, and Xiaojun Wan, editors,
  \emph{Proceedings of the 2019 Conference on Empirical Methods in Natural
  Language Processing and the 9th International Joint Conference on Natural
  Language Processing (EMNLP-IJCNLP)}, pages 4463--4473, Hong Kong, China,
  November 2019. Association for Computational Linguistics.
\newblock \doi{10.18653/v1/D19-1454}.

\bibitem[{Seed} et~al.(2025){Seed}, Li, Wu, Ma, Xiao, Zhang, Li, Wang, Xu,
  Zhang, Hu, Zan, Zhu, Xu, Li, Wu, Xia, Zhang, Shi, Chen, Zhu, Huang, Zhou,
  Dou, Duan, Lu, Jiang, Xu, Chen, Chen, Ma, Su, Chen, Wang, Yuan, Liu, Zhou,
  Hua, Shen, Xiang, Chen, Liu, Shen, Xiang, Yan, Luo, Zhang, Ding, Zhang,
  Liang, Li, Huang, Mu, Huang, Ma, Min, Yu, Pang, Zhang, Yan, Yan, Zhao, Cao,
  Wu, Chen, Li, Qiao, Sun, Xin, Fan, Huang, Fan, Jia, Zhang, Liu, Wu, Zuo, Jia,
  Yang, Liu, Yu, Bin, Hao, Luo, Li, Zhou, Peng, Chen, Lin, Leng, Li, Song, Ma,
  Shan, Xiang, Wu, Zhang, Yao, Bao, Yang, Yuan, Li, Xian, Zeng, Wang, Hong,
  Wang, Wang, Yang, Yin, Lu, Zhang, Chen, Zhang, Lin, Huang, Xu, Wei, and
  Wang]{seed2025vwn}
{Seed}, Baisheng Li, Banggu Wu, Bole Ma, Bowen Xiao, Chaoyi Zhang, Cheng Li,
  Chengyi Wang, Chengyin Xu, Chi Zhang, Chong Hu, Daoguang Zan, Defa Zhu,
  Dongyu Xu, Du~Li, Faming Wu, Fan Xia, Ge~Zhang, Guang Shi, Haobin Chen,
  Hongyu Zhu, Hongzhi Huang, Huan Zhou, Huanzhang Dou, Jianhui Duan, Jianqiao
  Lu, Jianyu Jiang, Jiayi Xu, Jiecao Chen, Jin Chen, Jin Ma, Jing Su, Jingji
  Chen, Jun Wang, Jun Yuan, Juncai Liu, Jundong Zhou, Kai Hua, Kai Shen, Kai
  Xiang, Kaiyuan Chen, Kang Liu, Ke~Shen, Liang Xiang, Lin Yan, Lishu Luo,
  Mengyao Zhang, Ming Ding, Mofan Zhang, Nianning Liang, Peng Li, Penghao
  Huang, Pengpeng Mu, Qi~Huang, Qianli Ma, Qiyang Min, Qiying Yu, Renming Pang,
  Ru~Zhang, Shen Yan, Shen Yan, Shixiong Zhao, Shuaishuai Cao, Shuang Wu, Siyan
  Chen, Siyu Li, Siyuan Qiao, Tao Sun, Tian Xin, Tiantian Fan, Ting Huang,
  Ting-Han Fan, Wei Jia, Wenqiang Zhang, Wenxuan Liu, Xiangzhong Wu, Xiaochen
  Zuo, Xiaoying Jia, Ximing Yang, Xin Liu, Xin Yu, Xingyan Bin, Xintong Hao,
  Xiongcai Luo, Xujing Li, Xun Zhou, Yanghua Peng, Yangrui Chen, Yi~Lin,
  Yichong Leng, Yinghao Li, Yingshuan Song, Yiyuan Ma, Yong Shan, Yongan Xiang,
  Yonghui Wu, Yongtao Zhang, Yongzhen Yao, Yu~Bao, Yuehang Yang, Yufeng Yuan,
  Yunshui Li, Yuqiao Xian, Yutao Zeng, Yuxuan Wang, Zehua Hong, Zehua Wang,
  Zengzhi Wang, Zeyu Yang, Zhengqiang Yin, Zhenyi Lu, Zhexi Zhang, Zhi Chen,
  Zhi Zhang, Zhiqi Lin, Zihao Huang, Zilin Xu, Ziyun Wei, and Zuo Wang.
\newblock Virtual width networks, 2025.

\bibitem[Shazeer(2020)]{shazeer2020glu}
Noam Shazeer.
\newblock Glu variants improve transformer, 2020.

\bibitem[Shazeer et~al.(2017)Shazeer, Mirhoseini, Maziarz, Davis, Le, Hinton,
  and Dean]{shazeer2017moe}
Noam Shazeer, Azalia Mirhoseini, Krzysztof Maziarz, Andy Davis, Quoc Le,
  Geoffrey Hinton, and Jeff Dean.
\newblock Outrageously large neural networks: The sparsely-gated
  mixture-of-experts layer.
\newblock In \emph{International Conference on Learning Representations}, 2017.

\bibitem[Shleifer et~al.(2021)Shleifer, Weston, and
  Ott]{shleifer2021normformer}
Sam Shleifer, Jason Weston, and Myle Ott.
\newblock Normformer: Improved transformer pretraining with extra
  normalization, 2021.

\bibitem[Soldaini et~al.(2024)Soldaini, Kinney, Bhagia, Schwenk, Atkinson,
  Authur, Bogin, Chandu, Dumas, Elazar, Hofmann, Jha, Kumar, Lucy, Lyu,
  Lambert, Magnusson, Morrison, Muennighoff, Naik, Nam, Peters, Ravichander,
  Richardson, Shen, Strubell, Subramani, Tafjord, Walsh, Zettlemoyer, Smith,
  Hajishirzi, Beltagy, Groeneveld, Dodge, and Lo]{soldaini2024dolma}
Luca Soldaini, Rodney Kinney, Akshita Bhagia, Dustin Schwenk, David Atkinson,
  Russell Authur, Ben Bogin, Khyathi Chandu, Jennifer Dumas, Yanai Elazar,
  Valentin Hofmann, Ananya Jha, Sachin Kumar, Li~Lucy, Xinxi Lyu, Nathan
  Lambert, Ian Magnusson, Jacob Morrison, Niklas Muennighoff, Aakanksha Naik,
  Crystal Nam, Matthew Peters, Abhilasha Ravichander, Kyle Richardson, Zejiang
  Shen, Emma Strubell, Nishant Subramani, Oyvind Tafjord, Evan Walsh, Luke
  Zettlemoyer, Noah Smith, Hannaneh Hajishirzi, Iz~Beltagy, Dirk Groeneveld,
  Jesse Dodge, and Kyle Lo.
\newblock Dolma: an open corpus of three trillion tokens for language model
  pretraining research.
\newblock In Lun-Wei Ku, Andre Martins, and Vivek Srikumar, editors,
  \emph{Proceedings of the 62nd Annual Meeting of the Association for
  Computational Linguistics (Volume 1: Long Papers)}, pages 15725--15788,
  Bangkok, Thailand, August 2024. Association for Computational Linguistics.
\newblock \doi{10.18653/v1/2024.acl-long.840}.

\bibitem[Srivastava et~al.(2015)Srivastava, Greff, and
  Schmidhuber]{srivastava2015highway}
Rupesh~Kumar Srivastava, Klaus Greff, and J{\"u}rgen Schmidhuber.
\newblock Highway networks, 2015.

\bibitem[Su et~al.(2024)Su, Ahmed, Lu, Pan, Bo, and Liu]{su2024roformer}
Jianlin Su, Murtadha Ahmed, Yu~Lu, Shengfeng Pan, Wen Bo, and Yunfeng Liu.
\newblock Roformer: Enhanced transformer with rotary position embedding.
\newblock \emph{Neurocomputing}, 568:\penalty0 127063, February 2024.
\newblock ISSN 0925-2312.
\newblock \doi{10.1016/j.neucom.2023.127063}.

\bibitem[Talmor et~al.(2019)Talmor, Herzig, Lourie, and
  Berant]{talmor2019commonsenseqa}
Alon Talmor, Jonathan Herzig, Nicholas Lourie, and Jonathan Berant.
\newblock {C}ommonsense{QA}: A question answering challenge targeting
  commonsense knowledge.
\newblock In Jill Burstein, Christy Doran, and Thamar Solorio, editors,
  \emph{Proceedings of the 2019 Conference of the North {A}merican Chapter of
  the Association for Computational Linguistics: Human Language Technologies,
  Volume 1 (Long and Short Papers)}, pages 4149--4158, Minneapolis, Minnesota,
  June 2019. Association for Computational Linguistics.
\newblock \doi{10.18653/v1/N19-1421}.

\bibitem[Touvron et~al.(2021)Touvron, Cord, Sablayrolles, Synnaeve, and
  J\'egou]{touvron2021layerscale}
Hugo Touvron, Matthieu Cord, Alexandre Sablayrolles, Gabriel Synnaeve, and
  Herv\'e J\'egou.
\newblock Going deeper with image transformers.
\newblock In \emph{Proceedings of the IEEE/CVF International Conference on
  Computer Vision (ICCV)}, pages 32--42, October 2021.

\bibitem[Vaswani et~al.(2017)Vaswani, Shazeer, Parmar, Uszkoreit, Jones, Gomez,
  Kaiser, and Polosukhin]{vaswani2017attention}
Ashish Vaswani, Noam Shazeer, Niki Parmar, Jakob Uszkoreit, Llion Jones,
  Aidan~N Gomez, \L~ukasz Kaiser, and Illia Polosukhin.
\newblock Attention is all you need.
\newblock In I.~Guyon, U.~Von Luxburg, S.~Bengio, H.~Wallach, R.~Fergus,
  S.~Vishwanathan, and R.~Garnett, editors, \emph{Advances in Neural
  Information Processing Systems}, volume~30. Curran Associates, Inc., 2017.

\bibitem[Walsh et~al.(2025)Walsh, Soldaini, Groeneveld, Lo, Arora, Bhagia, Gu,
  Huang, Jordan, Lambert, Schwenk, Tafjord, Anderson, Atkinson, Brahman, Clark,
  Dasigi, Dziri, Ettinger, Guerquin, Heineman, Ivison, Koh, Liu, Malik,
  Merrill, Miranda, Morrison, Murray, Nam, Poznanski, Pyatkin, Rangapur,
  Schmitz, Skjonsberg, Wadden, Wilhelm, Wilson, Zettlemoyer, Farhadi, Smith,
  and Hajishirzi]{olmo2025olmo2}
Evan~Pete Walsh, Luca Soldaini, Dirk Groeneveld, Kyle Lo, Shane Arora, Akshita
  Bhagia, Yuling Gu, Shengyi Huang, Matt Jordan, Nathan Lambert, Dustin
  Schwenk, Oyvind Tafjord, Taira Anderson, David Atkinson, Faeze Brahman,
  Christopher Clark, Pradeep Dasigi, Nouha Dziri, Allyson Ettinger, Michal
  Guerquin, David Heineman, Hamish Ivison, Pang~Wei Koh, Jiacheng Liu, Saumya
  Malik, William Merrill, Lester James~Validad Miranda, Jacob Morrison, Tyler
  Murray, Crystal Nam, Jake Poznanski, Valentina Pyatkin, Aman Rangapur,
  Michael Schmitz, Sam Skjonsberg, David Wadden, Christopher Wilhelm, Michael
  Wilson, Luke Zettlemoyer, Ali Farhadi, Noah~A. Smith, and Hannaneh
  Hajishirzi.
\newblock 2 {OLM}o 2 furious ({COLM}{\textquoteright}s version).
\newblock In \emph{Second Conference on Language Modeling}, 2025.

\bibitem[Wang et~al.(2024)Wang, Ma, Dong, Huang, Zhang, and
  Wei]{wang2022deepnet}
Hongyu Wang, Shuming Ma, Li~Dong, Shaohan Huang, Dongdong Zhang, and Furu Wei.
\newblock Deepnet: Scaling transformers to 1,000 layers.
\newblock \emph{IEEE Transactions on Pattern Analysis and Machine
  Intelligence}, 46\penalty0 (10):\penalty0 6761--6774, October 2024.
\newblock ISSN 1939-3539.
\newblock \doi{10.1109/tpami.2024.3386927}.

\bibitem[Wang et~al.(2019)Wang, Li, Xiao, Zhu, Li, Wong, and
  Chao]{wang2019learning}
Qiang Wang, Bei Li, Tong Xiao, Jingbo Zhu, Changliang Li, Derek~F. Wong, and
  Lidia~S. Chao.
\newblock Learning deep transformer models for machine translation.
\newblock In Anna Korhonen, David Traum, and Llu{\'i}s M{\`a}rquez, editors,
  \emph{Proceedings of the 57th Annual Meeting of the Association for
  Computational Linguistics}, pages 1810--1822, Florence, Italy, July 2019.
  Association for Computational Linguistics.
\newblock \doi{10.18653/v1/P19-1176}.

\bibitem[Welbl et~al.(2017)Welbl, Liu, and Gardner]{welbl2017sciq}
Johannes Welbl, Nelson~F. Liu, and Matt Gardner.
\newblock Crowdsourcing multiple choice science questions.
\newblock In Leon Derczynski, Wei Xu, Alan Ritter, and Tim Baldwin, editors,
  \emph{Proceedings of the 3rd Workshop on Noisy User-generated Text}, pages
  94--106, Copenhagen, Denmark, September 2017. Association for Computational
  Linguistics.
\newblock \doi{10.18653/v1/W17-4413}.

\bibitem[Xiao et~al.(2025)Xiao, Meng, Li, and Yuan]{xiao2025muddformer}
Da~Xiao, Qingye Meng, Shengping Li, and Xingyuan Yuan.
\newblock {MUDDF}ormer: Breaking residual bottlenecks in transformers via
  multiway dynamic dense connections.
\newblock In Aarti Singh, Maryam Fazel, Daniel Hsu, Simon Lacoste-Julien, Felix
  Berkenkamp, Tegan Maharaj, Kiri Wagstaff, and Jerry Zhu, editors,
  \emph{Proceedings of the 42nd International Conference on Machine Learning},
  volume 267 of \emph{Proceedings of Machine Learning Research}, pages
  68440--68458. PMLR, 13--19 Jul 2025.

\bibitem[Xie et~al.(2026)Xie, Wei, Cao, Zhao, Deng, Li, Dai, Gao, Xu, Yu, Zhao,
  Zhou, Xu, Zhang, Zeng, Hu, Wang, Yuan, Wang, and Liang]{xie2025mhc}
Zhenda Xie, Yixuan Wei, Huanqi Cao, Chenggang Zhao, Chengqi Deng, Jiashi Li,
  Damai Dai, Huazuo Gao, Mingyu Xu, Kuai Yu, Liang Zhao, Shangyan Zhou, Zhean
  Xu, Zhengyan Zhang, Wangding Zeng, Shengding Hu, Yuqing Wang, Jingyang Yuan,
  Lean Wang, and Wenfeng Liang.
\newblock m{HC}: Manifold-constrained hyper-connections.
\newblock In \emph{Forty-third International Conference on Machine Learning},
  2026.

\bibitem[Xiong et~al.(2020)Xiong, Yang, He, Zheng, Zheng, Xing, Zhang, Lan,
  Wang, and Liu]{xiong2020layer}
Ruibin Xiong, Yunchang Yang, Di~He, Kai Zheng, Shuxin Zheng, Chen Xing,
  Huishuai Zhang, Yanyan Lan, Liwei Wang, and Tieyan Liu.
\newblock On layer normalization in the transformer architecture.
\newblock In Hal~Daum{\'e} III and Aarti Singh, editors, \emph{Proceedings of
  the 37th International Conference on Machine Learning}, volume 119 of
  \emph{Proceedings of Machine Learning Research}, pages 10524--10533. PMLR,
  13--18 Jul 2020.

\bibitem[Zellers et~al.(2019)Zellers, Holtzman, Bisk, Farhadi, and
  Choi]{zellers2019hellaswag}
Rowan Zellers, Ari Holtzman, Yonatan Bisk, Ali Farhadi, and Yejin Choi.
\newblock {H}ella{S}wag: Can a machine really finish your sentence?
\newblock In Anna Korhonen, David Traum, and Llu{\'i}s M{\`a}rquez, editors,
  \emph{Proceedings of the 57th Annual Meeting of the Association for
  Computational Linguistics}, pages 4791--4800, Florence, Italy, July 2019.
  Association for Computational Linguistics.
\newblock \doi{10.18653/v1/P19-1472}.

\bibitem[Zhang and Sennrich(2019)]{zhang2019rmsnorm}
Biao Zhang and Rico Sennrich.
\newblock Root mean square layer normalization.
\newblock In H.~Wallach, H.~Larochelle, A.~Beygelzimer, F.~d\textquotesingle
  Alch\'{e}-Buc, E.~Fox, and R.~Garnett, editors, \emph{Advances in Neural
  Information Processing Systems}, volume~32. Curran Associates, Inc., 2019.

\bibitem[Zhang et~al.(2026)Zhang, Liu, Wang, and Gu]{zhang2026ddl}
Yifan Zhang, Yifeng Liu, Mengdi Wang, and Quanquan Gu.
\newblock Deep delta learning, 2026.

\bibitem[Zhou et~al.(2025)Zhou, Wu, Jiang, Obeid, and
  Lan]{zhou2025valueresidual}
Zhanchao Zhou, Tianyi Wu, Zhiyun Jiang, Fares Obeid, and Zhenzhong Lan.
\newblock Value residual learning.
\newblock In Wanxiang Che, Joyce Nabende, Ekaterina Shutova, and Mohammad~Taher
  Pilehvar, editors, \emph{Proceedings of the 63rd Annual Meeting of the
  Association for Computational Linguistics (Volume 1: Long Papers)}, pages
  28341--28356, Vienna, Austria, July 2025. Association for Computational
  Linguistics.
\newblock ISBN 979-8-89176-251-0.
\newblock \doi{10.18653/v1/2025.acl-long.1375}.

\bibitem[Zhu et~al.(2025{\natexlab{a}})Zhu, Huang, Huang, Zeng, Mao, Wu, Min,
  and Zhou]{zhu2025hyperconnections}
Defa Zhu, Hongzhi Huang, Zihao Huang, Yutao Zeng, Yunyao Mao, Banggu Wu, Qiyang
  Min, and Xun Zhou.
\newblock Hyper-connections.
\newblock In Y.~Yue, A.~Garg, N.~Peng, F.~Sha, and R.~Yu, editors,
  \emph{International Conference on Learning Representations}, volume 2025,
  pages 97183--97219, 2025{\natexlab{a}}.

\bibitem[Zhu et~al.(2025{\natexlab{b}})Zhu, Huang, Zhou, Huang, Zeng, Wu, Min,
  and Zhou]{zhu2025frac}
Defa Zhu, Hongzhi Huang, Jundong Zhou, Zihao Huang, Yutao Zeng, Banggu Wu,
  Qiyang Min, and Xun Zhou.
\newblock Frac-connections: Fractional extension of hyper-connections,
  2025{\natexlab{b}}.

\bibitem[Zhuo et~al.(2025)Zhuo, Zeng, Wang, Zhang, Yang, Li, Zhou, and
  Ma]{zhuo2025hybridnorm}
Zhijian Zhuo, Yutao Zeng, Ya~Wang, Sijun Zhang, Jian Yang, Xiaoqing Li, Xun
  Zhou, and Jinwen Ma.
\newblock Hybridnorm: Towards stable and efficient transformer training via
  hybrid normalization.
\newblock In D.~Belgrave, C.~Zhang, H.~Lin, R.~Pascanu, P.~Koniusz,
  M.~Ghassemi, and N.~Chen, editors, \emph{Advances in Neural Information
  Processing Systems}, volume 38, Main Conference, pages 25106--25147. Curran
  Associates, Inc., 2025.
\newblock \doi{10.52202/085713-0844}.

\end{thebibliography}
\end{document}